\documentclass[10pt,twocolumn,letterpaper]{article}

\PassOptionsToPackage{table}{xcolor}

\usepackage[pagenumbers]{cvpr} 
\usepackage{algorithm}
\usepackage{algorithmic}
\usepackage{multirow}
\usepackage{makecell}
\usepackage{booktabs}
\usepackage{appendix}
\definecolor{cvprblue}{rgb}{0.21,0.49,0.74}
 
\newcommand{\lingan}[1]{{\color{black}#1}}

\usepackage[pagebackref,breaklinks,colorlinks,allcolors=cvprblue]{hyperref}

\def\paperID{5277} 
\def\confName{CVPR}
\def\confYear{2025}

\title{ChainHOI: Joint-based Kinematic Chain Modeling for \\Human-Object Interaction Generation\vspace{-0.2cm}}

\author{Ling-An Zeng$^{\dagger}$, Guohong Huang$^\dagger$, Yi-Lin Wei, Shengbo Gu, \\ 
Yu-Ming Tang, Jingke Meng*, Wei-Shi Zheng*\\
Sun Yat-sen University, China \\
Key Laboratory of Machine Intelligence and Advanced Computing, Ministry of Education, China \\
{\tt\small \{zenglan3, huanggh37\}@mail2.sysu.edu.cn, mengjke@gmail.com, wszheng@ieee.org}
}

\begin{document}

\twocolumn[{%
\renewcommand\twocolumn[1][]{#1}%
\maketitle
\vspace{-1.25cm}
\begin{center}
    \centering
    \includegraphics[width=\linewidth]{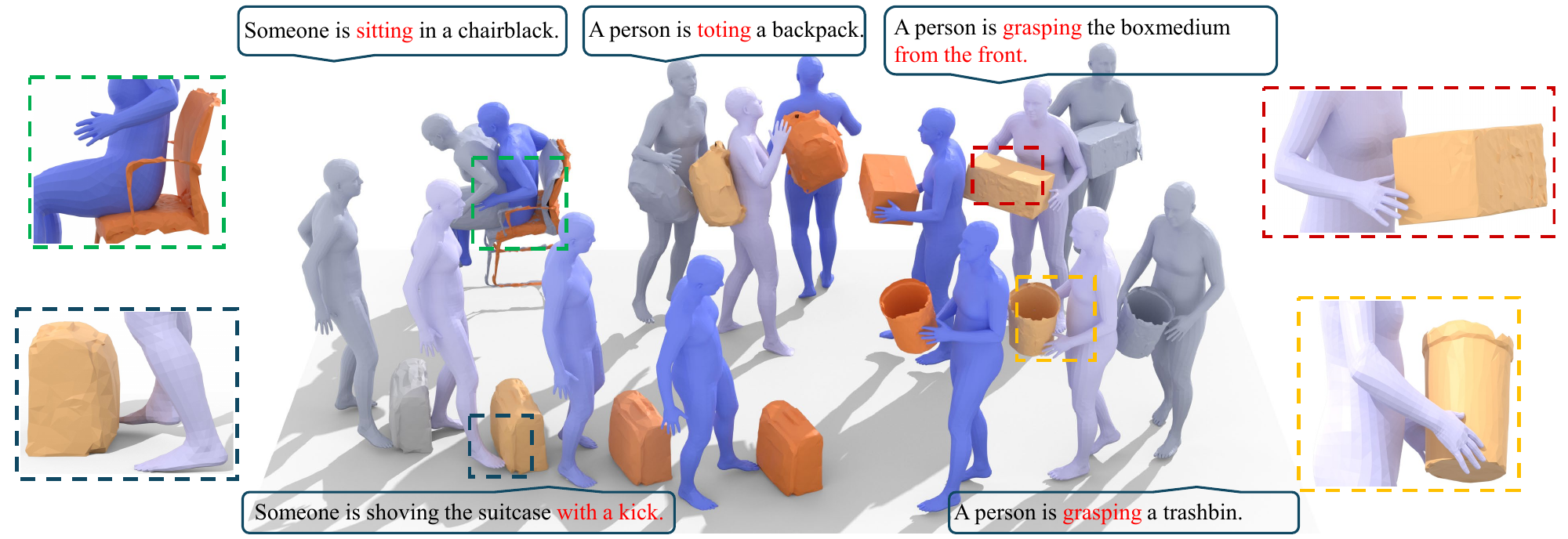}
    \vspace{-0.7cm}
    \captionof{figure}{Given a text description and target object geometry, our ChainHOI effectively generates high-quality human-object interaction sequences that are both logical and realistic.}
\end{center}%
}]

\def\thefootnote{}\footnotetext{$\dagger$ Equal contribution; * Corresponding authors.}


\vspace{-0.5cm}
\begin{abstract}
    We propose ChainHOI, a novel approach for text-driven human-object interaction (HOI) generation that explicitly models interactions at both the joint and kinetic chain levels.
    Unlike existing methods that implicitly model interactions using full-body poses as tokens, we argue that explicitly modeling joint-level interactions is more natural and effective for generating realistic HOIs, as it directly captures the geometric and semantic relationships between joints, rather than modeling interactions in the latent pose space.
    To this end, ChainHOI introduces a novel joint graph to capture potential interactions with objects, and a Generative Spatiotemporal Graph Convolution Network to explicitly model interactions at the joint level. 
    Furthermore, we propose a Kinematics-based Interaction Module that explicitly models interactions at the kinetic chain level, ensuring more realistic and biomechanically coherent motions.
    Evaluations on two public datasets demonstrate that ChainHOI significantly outperforms previous methods, generating more realistic, and semantically consistent HOIs. Code is available \href{https://github.com/qinghuannn/ChainHOI}{here}.
\end{abstract}

\vspace{-0.5cm}

\section{Introduction} \vspace{-0.1cm}
Text-driven human-object interaction (HOI) generation aims to generate realistic and semantically coherent motions for both humans and objects based on provided textual descriptions. Unlike previous approaches relying solely on object motions \cite{omomo} or past motions \cite{interdiff}, text-driven HOI generation offers greater controllability. This enhanced control holds significant value for applications in AR/VR, gaming, and film production \cite{cghoi, interdiff, hoianimator, thor, chois}.
 
\lingan{Since the advent of Transformer architectures \cite{transformer}, which have revolutionized the motion generation domain, existing HOI methods \cite{omomo, cghoi, interdiff, hoianimator, thor, chois} typically embed full-body poses as tokens within each frame, which are then processed by Transformers to generate human motion.} While this approach has proven effective, it models human skeletal joints implicitly, rather than capturing their interactions explicitly. 
Such implicit modeling increases the difficulty of model learning and lacks a clear understanding of physics and geometry in the explicit space.
\lingan{We argue that explicitly modeling joint-level interactions is more natural and effective for HOI generation, as it directly captures the geometric and semantic relationships between joints, rather than modeling interactions in the latent pose space.}

Moreover, the joint-level representation allows us to explicitly model HOIs at the kinetic chain level. We claim that explicit kinetic chain-level modeling is essential for generating realistic HOIs, as it captures the interdependencies and coordination between different body parts during motion, ensuring that generated actions adhere to biomechanical principles. For example, during a grasping motion, all the joints in the arm's kinetic chain—including the shoulder, elbow, and wrist—must work in unison to perform the action, while the joints in the leg's kinetic chain allow the body to assist in positioning the body relative to the object. Therefore, modeling HOIs based on kinetic chains can generate more natural and fluid interaction movements.

Motivated by \lingan{the above} considerations, this work proposes a method to explicitly model human-object interactions at both the joint and kinetic chain levels. To achieve joint-level interaction modeling, we introduce a novel joint graph that captures potential interactions with objects and develop a \textbf{Generative Spatiotemporal Graph Convolution Network} (GST-GCN) \lingan{to model joint relations}. Specifically, we incorporate an object node to capture information beyond the human skeletal joints and establish edges only between the object node and the potential interaction joints. \lingan{These selective connections} enhance the network's effectiveness and precision in modeling HOIs. The GST-GCN comprises a well-designed spatiotemporal graph convolution network for capturing short-term interactions and a semantic-consistent module for modeling long-term information while maintaining semantic consistency. This architecture enables the generation of semantically coherent long HOI sequences conditioned on text, while explicitly modeling interactions between different nodes.

Considering the benefits of kinetic chain-level interaction modeling mentioned above, we introduce a specialized kinetic chain framework for HOI and propose a \textbf{Kinematics-based Interaction Module} (KIM) to capture interactions at the kinetic chain level. Similar to the joint graph design, we incorporate an additional kinetic chain that connects the object node with potential interacting human joints, thereby representing potential interactions. Inspired by DETR \cite{detr}, our KIM leverages Transformer decoders and incorporates learnable kinetic chain tokens to model specific kinetic chains. A dedicated attention mask ensures that each token attends only to nodes within its corresponding kinetic chain. Additionally, KIM employs a secondary Transformer decoder, where the keys and values are derived from text and object geometry, enabling kinetic chain tokens to encode specific interaction intents and object geometry information. Thus, our KIM provides explicit and detailed modeling of interactions at the kinetic chain level, \lingan{enhancing the precision of generated HOIs.}

By integrating the aforementioned designs, we introduce a novel method named \textbf{ChainHOI} for text-driven HOI generation that explicitly models interactions at both the joint and kinetic chain levels. This dual-level approach ensures more natural, realistic, and semantically consistent human-object interactions. We evaluate ChainHOI on the public BEHAVE \cite{behave} and OMOMO \cite{omomo} datasets. Extensive experimental results demonstrate the effectiveness of ChainHOI, highlighting the superiority of explicitly modeling interactions at both joint and kinetic chain levels.

In summary, our main contributions are as follows:
\begin{itemize}
    \item We propose \textbf{ChainHOI}, a novel method that explicitly models interactions at both joint and kinetic chain levels, demonstrating the effectiveness of this dual-level design for text-driven HOI generation. 
    \item We introduce a novel joint graph for HOI and develop a \textbf{Generative Spatiotemporal Graph Convolution Network} (GST-GCN), which effectively models both short-term and long-term information interactions \textbf{at the joint level} while ensuring semantic consistency. 
    \item We design an effective \textbf{Kinematics-based Interaction Module} (KIM) that provides explicit and comprehensive modeling of interactions \textbf{at the kinetic chain level}, enhancing the realism and coherence of generated human-object interactions.
\end{itemize}

\section{Related Work}
\noindent\textbf{Human-Object Interaction Generation.}
Human-object interaction generation is an emerging research field, and many works explore different settings. \lingan{Some studies employ reinforcement learning to develop physics-based methods \cite{hassan2023synthesizing, xie2023hierarchical, merel2020catch}, though these models often struggle to generalize to diverse objects and are difficult to train.} Other studies utilize generative models to produce HOIs. Few works focus on hand-object interactions \cite{text2hoi, manipnet, cams, christen2022d}. Advancements in human motion generation have led more researchers to explore full-body motion. \lingan{For example, OMOMO \cite{omomo} generates full-body motion conditioned on object motions, while InterDiff \cite{interdiff} predicts both human and object motions. To enhance controllability, methods such as HOI-Diff \cite{hoidiff}, CG-HOI \cite{cghoi}, and HOIAnimator \cite{hoianimator} generate human and object motions based on text.} Additionally, CHOIS \cite{chois} generates motions guided by the initial states of humans and objects, as well as by text and object waypoints. \lingan{In this work, our method is based on generative models and generates both full-body human and object motions conditioned on text.} In contrast to existing HOI methods, our approach explicitly models human-object interactions at both the joint and kinetic chain levels.

\begin{figure*}[!ht]
    \centering
    \includegraphics[width=\linewidth]{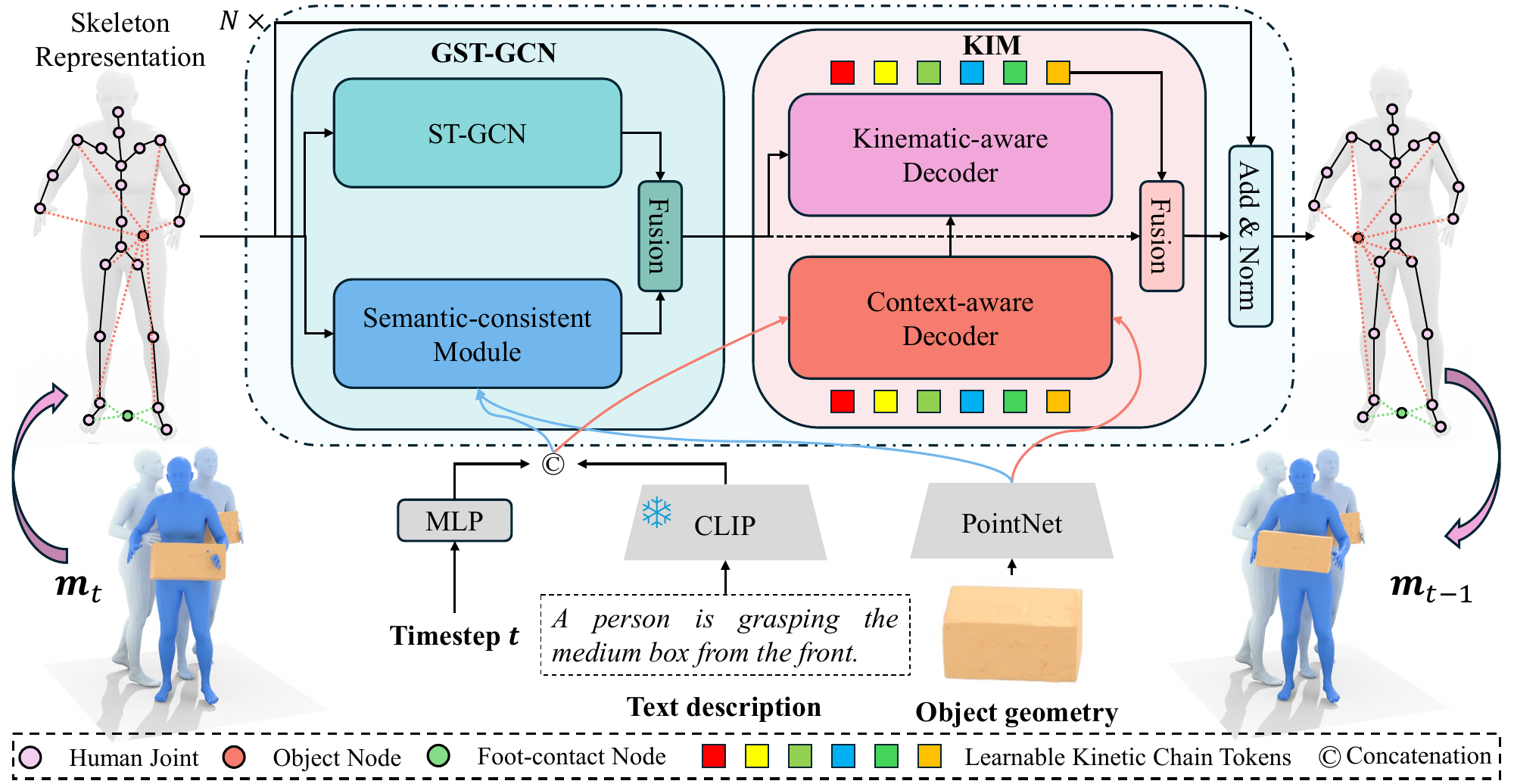}
    \vspace{-0.6cm}
    \caption{\textbf{Overview of ChainHOI.} ChainHOI is a diffusion-based model with $N$ identical blocks. Each block contains a Generative Spatiotemporal GCN (GST-GCN) and a Kinematics-based Interaction Module (KIM) to model interactions at the joint and kinetic chain levels. GST-GCN, comprising an ST-GCN and a Semantic-consistent Module, captures short- and long-term information while ensuring semantic consistency. KIM includes a Context-aware Decoder and a Kinematic-aware Decoder to capture HOI context (textual and object geometry) and to model intra- and inter-kinetic chain interactions. Input and output projection layers are omitted for clarity.}
    \label{fig:overview}
\end{figure*}

\vspace{0.15cm}
\noindent\textbf{Human-Scene Motion Generation.}
Human-scene motion generation focuses on perceiving the environment and models how dynamic human movements engage with static surroundings. Existing methods explore unconditional motion generation \cite{liu2023revisit, Synthesizing}, text-guided motion generation \cite{jiang2024scaling, yi2024generating, vuong2024language, xiao2023unified, wang2022humanise}, interaction-field-guided motion generation \cite{nifty}, and motion forecasting \cite{Humans}. Unlike these approaches, which are limited to static scenes, this work investigates interactions between humans and dynamic objects.

\vspace{0.15cm}
\noindent\textbf{Human Motion Generation.}
Most research has focused on conditional full-body motion generation, with limited exploration of unsupervised human motion \cite{raab2023modi, holden2016deep}. Several studies address music-guided \cite{le2023music, tseng2023edge, sun2022you, wang2022groupdancer, au2022choreograph, li2024techcoach, zeng2024multimodal, lin2024human} and speech-guided motion generation \cite{liu2022learning, liu2022disco, liu2022beat, habibie2022motion, ao2022rhythmic, zhu2023taming, yi2023generating, alexanderson2023listen, ao2023gesturediffuclip}. For text-driven motion generation, studies have explored various models, including diffusion-based models \cite{mld, attt2m, md, fgt2m, mdm, remodiffuse, mofusion, motionlcm, emdm, CrossDiff, diversemotion, bridgegap, jin2024local, sampieri2024length, light-t2m}, autoregressive models \cite{motiongpt, t2m-gpt, avatargpt, parco, humantomato}, generative masked models \cite{momask, mmm, bamm}, and VAE-based models \cite{temos, t2m}. Several works also focus on more controllable motion generation \cite{gmd, omnicontrol, priorMDM, tlcontrol, wandr}. In this work, we aim to generate both human and object motions, not just human motion.

\section{Method}


\subsection{Preliminarily}
\label{sec: pre}
\noindent\textbf{Text-driven Human-object Interaction Definition.}
Given a text description $C$ and static object geometry $G$, the goal is to generate a 3D HOI sequence $\mathbf{m}$ of length $L$. We propose a new joint-level HOI representation method that maintains the advantages of redundant human motion representations widely used in text-driven human motion generation \cite{t2m}, while adding global information of the root (pelvis) joint and 6-DoF information of the object. 

For the root joint, our representation includes angular velocity $r^a \in \mathbb{R}$ along the Y-axis, linear velocities $r^v \in \mathbb{R}^2$ on the XZ-plane, height $r^y \in \mathbb{R}$, and global positions $r^p \in \mathbb{R}^3$. For non-root joints, the $D_{in}$-dimensional vector contains local joint positions $j^p \in \mathbb{R}^3$, velocities $j^v \in \mathbb{R}^3$, and rotations $j^r \in \mathbb{R}^6$ in root space. Additionally, an extra virtual joint includes binary foot-ground contact features $c^f \in \mathbb{R}^4$. For the object, the object's 6-DoF information contains the global rotation and translation. The representations of the virtual joint and object node are zero-padded to $D_{in}$ dimensions ($D_{in}=12$). In this way, a 3D HOI sequence is represented by a vector $\mathbf{m} \in \mathbb{R}^{L \times (J+2) \times D_{in}}$.

\vspace{0.15cm}
\noindent\textbf{Diffusion Models.}
We adopt a diffusion-based model for generating human-object interactions, leveraging the strengths of denoising diffusion probabilistic models \cite{ho2020denoising}. In the forward process, Gaussian noise is progressively added to the data until reaching pure noise at timestep $T$:
\begin{equation}
q(\mathbf{m}_t \mid \mathbf{m}_0) = \mathcal{N}(\mathbf{m}_t; \sqrt{\bar{\alpha}_t} \mathbf{m}_0, (1 - \bar{\alpha}_t) \mathbf{I}),
\end{equation}
where $\bar{\alpha}_t = \prod_{s=1}^t (1 - \beta_s)$ is the cumulative noise schedule. The reverse process is modeled by a neural network that iteratively denoises the data, thereby generating realistic human-object interactions.

\begin{figure}[t]
    \centering
    \includegraphics[width=\linewidth]{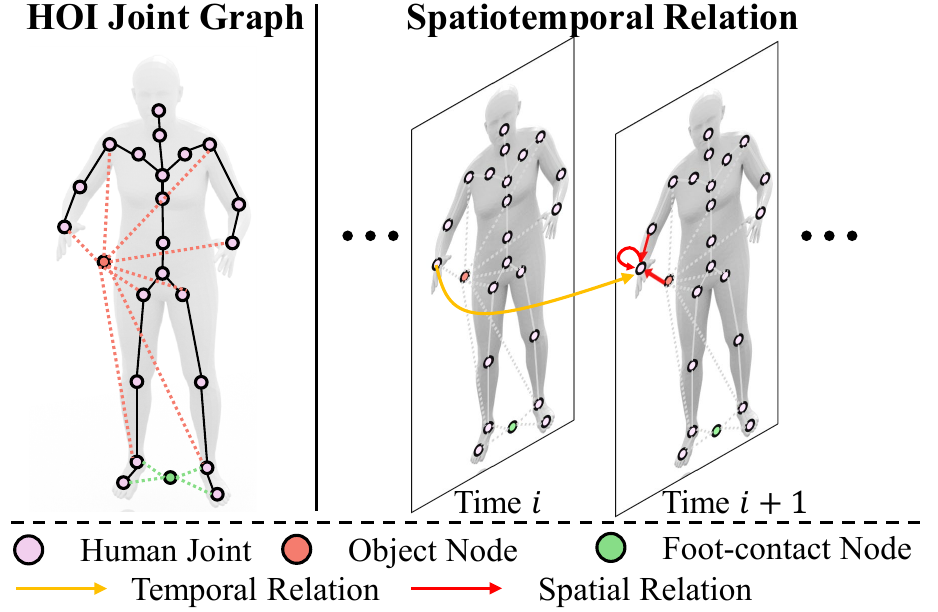}
    \vspace{-0.6cm}
    \caption{\textbf{Design of the HOI Joint Graph.} The object node contains object information and is connected to potential interaction joints. The foot-contact node is added to prevent foot sliding.}
    \label{fig:graph}
\end{figure}

\subsection{Overview of Our ChainHOI}
An overview of our method is shown in \cref{fig:overview}. ChainHOI is a diffusion-based approach in which the noisy 3D HOI sequence $\mathbf{m}_t$ is initially represented at the joint level, as described in Section \ref{sec: pre}. We employ PointNet \cite{pointnet} and a fixed pretrained CLIP \cite{clip} to extract object geometry and text embedding features, respectively. The text and object geometry features are projected into the $D_t$ dimension via linear layers. The HOI inputs are projected into the $D_m$ dimension via another linear layer. All inputs are then fed into ChainHOI to predict the denoised output $\mathbf{m}_{t-1}$.

Specifically, ChainHOI consists of $N$ identical blocks, each comprising a Generative Spatiotemporal Graph Convolution Network (GST-GCN) and a Kinematics-based Interaction Module (KIM). The GST-GCN leverages ST-GCN \cite{stgcn, stgcn++} and a Semantic-consistent Module to model both short-term and long-term information interactions at the joint level while ensuring semantic consistency. For the KIM, a set of learnable kinetic chain tokens is used to represent different kinematic chains. These tokens are first fed into a Context-aware Decoder to capture the HOI goal and object geometry. Subsequently, a Kinematic-aware Decoder processes the context-aware tokens and the outputs from the GST-GCN to model interactions at the kinetic chain level. Finally, the obtained tokens and the GST-GCN outputs are fused via a Fusion Module.

\begin{figure}[t]
    \centering
    \includegraphics[width=\linewidth]{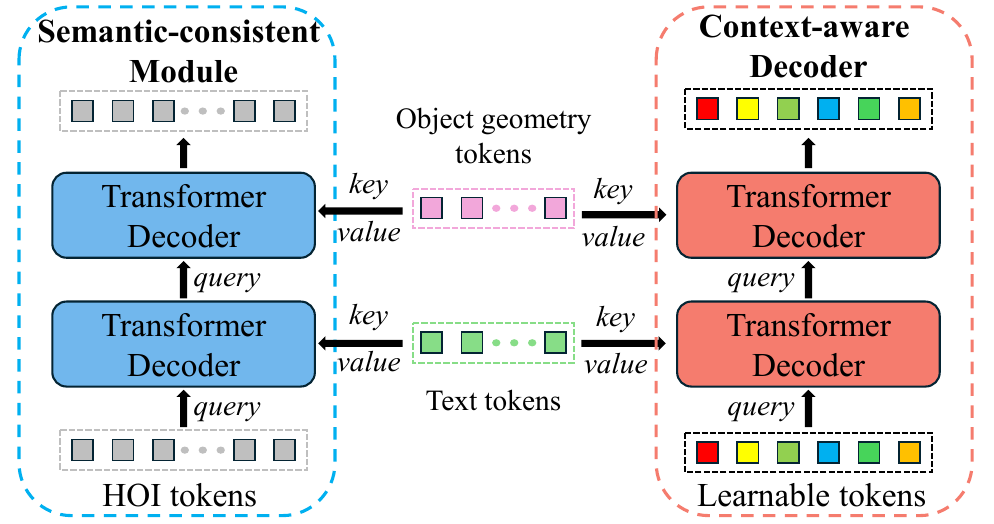}
    \vspace{-0.6cm}
    \caption{\textbf{Semantic-consistent Module and Context-aware Decoder.} Due to input differences, both modules have a similar structure, though their objectives differ. The former models long-term information and ensures semantic consistency, while the latter models context to plan the goals of each kinetic chain.}
    
    \label{fig:decoder}
\end{figure}

\subsection{Joint Level-Interaction Modeling}
To explicitly model interactions at the joint level, \lingan{Our GST-GCN introduces} a novel HOI joint graph that represents the connections between human joints and object nodes. Given the module inputs $\mathbf{z} \in \mathbb{R}^{L \times (J+2) \times D_m}$, we adopt the ST-GCN \cite{stgcn, stgcn++} to capture the short-term information between connected nodes. Since the ST-GCN cannot model long-term information or facilitate conditionally guided generation, we design a Semantic-consistent Module to capture long-term information and inject conditional guidance during generation. Finally, we fuse the outputs of the ST-GCN and the Semantic-consistent Module.

\vspace{0.15cm}
\noindent\textbf{Design of HOI Joint Graph.}
As shown in \cref{fig:graph}, to design the graph for the HOI generation task, we introduce an object node that encapsulates the information of the object. To capture potential interactions between the object and human joints, we add multiple edges. Following HOI-Diff \cite{hoidiff}, we consider only eight \lingan{potential interaction} human joints: the pelvis, neck, feet, shoulders, and hands. Additionally, to prevent foot sliding, we add a foot-contact node and establish edges between it and the foot joints.

\vspace{0.15cm}
\noindent\textbf{Modeling Short-term Information.}
Building on the success of graph convolutional networks in action recognition, we adopt the well-established ST-GCN \cite{stgcn, stgcn++} to capture short-term spatiotemporal information between connected nodes. Specifically, our ST-GCN module comprises a spatial graph convolution layer and a multi-branch temporal convolution layer. For more details, please refer to \cite{stgcn, stgcn++}.

\vspace{0.15cm}
\noindent\textbf{Modeling Long-term Information and Semantics.}
As shown in the left part of \cref{fig:decoder}, the Semantic-consistent Module uses two decoders to capture text information and object geometry to model long-term information. The reshaped input tensor $\mathbf{z} \in \mathbb{R}^{L \times ((J+2) \times D_m)}$ is first converted to HOI tokens $\bar{\mathbf{z}} \in \mathbb{R}^{L \times D_t}$ via a linear layer and average pooling. During each decoding step, the HOI tokens serve as \textit{queries}, and the object geometry and text tokens serve as \textit{keys} and \textit{values}, respectively.

\vspace{0.15cm}
\noindent\textbf{Fusing Outputs.}
To integrate the outputs of the ST-GCN and the Semantic-consistent Module ($\mathbf{z}^s$ and $\mathbf{z}^l$), we unsqueeze and repeat $\mathbf{z}^l$ along the time dimension and apply a linear layer to map its dimension from $D_t$ to $D_m$. Then, $\mathbf{z}^l \in \mathbb{R}^{L \times (J+2) \times D_m}$ and $\mathbf{z}^s \in \mathbb{R}^{L \times (J+2) \times D_m}$ are concatenated and passed through another linear layer to project them back to $D_m$:
\begin{equation} 
    \mathbf{y} = \text{Linear}([\mathbf{z}^l ; \; \mathbf{z}^s]). 
\end{equation}

\vspace{-0.1cm}

\begin{figure}
    \centering
    \includegraphics[width=\linewidth]{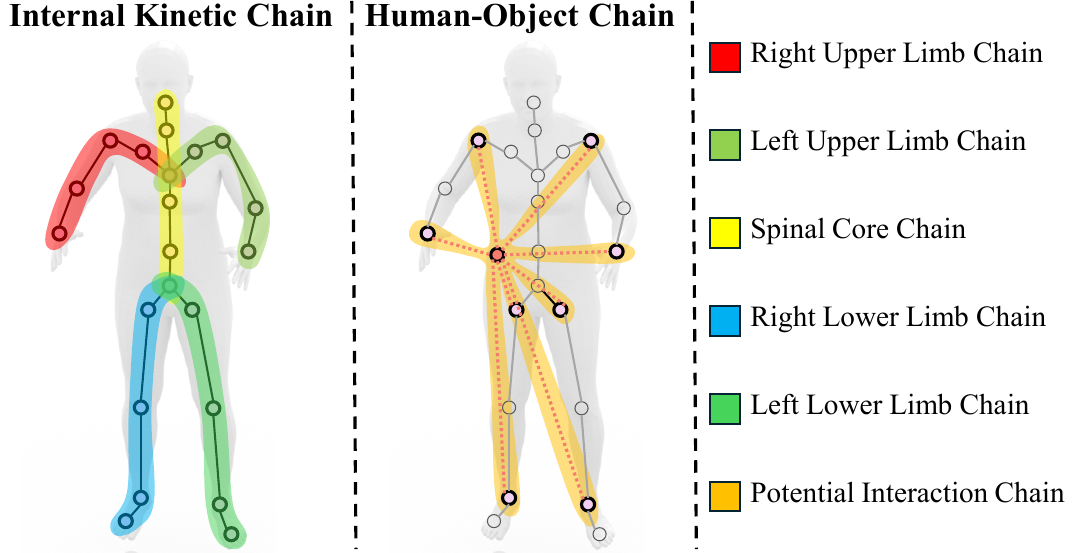}
    \vspace{-0.4cm}
    \caption{\textbf{Design of Kinetic Chains.} Beyond internal kinetic chains, an additional interaction chain is used to explicitly model the interactions between joints and the object.}
    \vspace{-0.2cm}
    \label{fig:chain}
\end{figure}

\subsection{Kinetic Chain-Level Interaction Modeling}
To explicitly model interactions at the kinetic chain level, we design a novel HOI kinetic chain framework based on our proposed HOI joint graph, to capture complex movement coordination. We employ learnable kinetic chain tokens $\mathbf{v}$ to represent specific kinetic chains. Since planning the goal of each kinetic chain is not possible without the HOI target and object geometry, we first input these tokens into a Context-aware Decoder to acquire contextual information. Then, the resulting tokens and outputs of the GST-GCN $\mathbf{y}$ are passed to the Kinematic-aware Decoder to capture inter- and intra-kinetic chain interactions. Finally, the Fusion Module integrates the obtained tokens $\mathbf{v}^\prime$ with $\mathbf{y}$.

\vspace{0.15cm}
\noindent\textbf{Design of Kinetic Chains.}
As shown in \cref{fig:chain}, our HOI kinetic chain framework includes five internal chains and one human-object chain. The human-object chain consists of the object node and eight potential interaction joints. These edges allow our KIM to explicitly model the interactions between the object and human joints.

\vspace{0.15cm}
\noindent\textbf{Capturing Context Information.}
As shown on the right side of \cref{fig:decoder}, the structure of the Context-aware Decoder is identical to that of the Semantic-consistent Module, where the \textit{queries} are learnable tokens, in contrast to HOI tokens.

\vspace{0.15cm}
\noindent\textbf{Intra- and Inter-Kinetic Chain Modeling.}
The Kinematic-aware Decoder is also built upon the Transformer Decoder and focuses on kinematic modeling within each frame. For the $i$-th frame, $\mathbf{y}_i \in \mathbb{R}^{(J+2) \times D_m}$ is projected into a $D_t$-dimensional space (denoted as $\bar{\mathbf{y}}_i \in \mathbb{R}^{(J+2) \times D_t}$) via a linear layer, and then serves as \textit{keys} and \textit{values}. The learnable tokens refined by the Context-aware Decoder serve as \textit{queries}. Formally, denote joint tokens as JT and learnable kinetic chain tokens as KT. Intra- and inter-chain modeling are formulated as:
\begin{align}
    \text{KT}' \, \text{=} \, \allowbreak \text{SelfAtt}( \allowbreak q\text{=}KT, \allowbreak k\text{=}KT, \allowbreak v\text{=}KT), \\
    \bar{\text{KT}} \, \text{=} \, \text{CrossAtt}( \allowbreak q\text{=}KT', \allowbreak  k\text{=}JT, \allowbreak v\text{=}JT, \text{mask}\text{=}M).
\end{align}
In $\text{SelfAtt}()$, KTs exchange information among themselves for inter-kinetic chain modeling; in $\text{CrossAtt}()$, M ensures that each KT can only attend to JTs in the corresponding kinetic chain, thus achieving intra-kinetic chain modeling.


\vspace{0.15cm}
\noindent\textbf{Fusing Kinetic Chain Tokens and Inputs.}
We design a fusion module to integrate the processed inputs $\bar{\mathbf{y}}_i \in \mathbb{R}^{(J+2) \times D_t}$ with the obtained KT $\mathbf{v}^\prime \in \mathbb{R}^{6 \times D_t}$ (6 denotes the number of chains) that contain kinetic chain information. We flatten $\mathbf{v}'$ ($\mathbb{R}^{6 D_t}$), pass it through a fully connected layer, and reshape the output from $\mathbb{R}^{(J+2)D_t}$ to $\mathbb{R}^{(J+2)\times D_t}$, yielding $\mathbf{v}''$.
We then concatenate $\mathbf{v}''$ with the module inputs $\bar{\mathbf{y}}_i$ and use a linear layer to project them back to the original dimension $D_m$:
\begin{equation}
 \hat{\mathbf{y}}_i = \text{Linear}([\mathbf{v}''; \;\bar{\mathbf{y}}_i]).
\end{equation}

\subsection{Training Losses}
To enhance HOI sequence quality, we introduce two auxiliary losses addressing the diffusion loss's lack of explicit contact constraints. \lingan{To avoid penetration,} an ideal approach would be to use the Signed Distance Function (SDF) between the human mesh and the object mesh as the loss. However, the computation of this loss is extremely slow due to the time-consuming conversion from 3D motion to the human mesh and the large number of human meshes. Thus, we propose a simplified loss function. For the human body, because of the limited number of joints, we calculate the distances between the \lingan{eight predicted potential interaction} joints and the GT object mesh:
\begin{equation}
    \mathcal{L}_h = \sum_{i=1}^{L} \sum_{k=1}^{8} a_{i,k} \cdot \; \mathcal{G}(\phi_h (H_{i,k}),\; \psi (O^{gt}_i)),
\end{equation}
where $\phi_h (H_{i,k})$ and $\psi (O^{gt}_i)$ denote the absolute positions of joint $k$ and the GT object mesh in the $i$-th frame. $\mathcal{G}()$ computes the square of the minimum absolute distance from the joint to all triangles. The contact label $a_{i,k}$, indicating whether joint $k$ is in contact with the object in the $i$-th frame, is preprocessed based on the GT HOI sequence.


\begin{table*}[h]
\centering
\setlength\tabcolsep{1mm}
\begin{tabular}{lcccccccc}
\bottomrule
 \multirow{2}{*}{Methods} & \multirow{2}{*}{AIT$\downarrow$} & \multirow{2}{*}{FID$\downarrow$}  & \multicolumn{3}{c}{R-Precision$\uparrow$} & \multirow{2}{*}{OCD $\downarrow$} & \multirow{2}{*}{PS $\downarrow$} & \multirow{2}{*}{FSR $\downarrow$} \\ \cline{4-6}
   & & & Top1 & Top2 & Top3 & & \\ \toprule
\rowcolor{gray!30} \multicolumn{9}{l}{\textit{On the BEHAVE dataset}}\\ \hline 
MDM$^{finetuned}$ \cite{mdm} & $5.31$s  & $0.246^{\pm.006}$ & $0.223^{\pm.011}$ & $0.378^{\pm.105}$ & $0.488^{\pm.015}$ & - & - & -  \\
MDM$^\star$ \cite{mdm} &$5.34$s & $0.257^{\pm.004}$ & $0.220^{\pm.007}$ & $0.355^{\pm.001}$ & $0.451^{\pm.001}$ & $0.458^{\pm.013}$ & $0.095^{\pm.007}$ & $0.098^{\pm.002}$\\
PriorMDM$^\star$ \cite{priorMDM} & $38.4$s & $0.328^{\pm.018}$ & $0.243^{\pm.009}$ & $0.329^{\pm.009}$ & $0.385^{\pm.013}$ & $0.215^{\pm.012}$ & $0.116^{\pm.001}$ & $0.066^{\pm.004}$\\
InerDiff \cite{interdiff} & $16.5$s & $0.170^{\pm.002}$ & $0.310^{\pm.003}$ & $0.480^{\pm.005}$ & $0.599^{\pm.001}$ & $0.191^{\pm.027}$ & $\pmb{0.078}^{\pm.000}$ & $0.069^{\pm.002}$\\
CHOIS$^\star$ \cite{chois} & - & $0.157^{\pm.001}$ & $0.301^{\pm.002}$ & $0.488^{\pm.003}$ & $0.606^{\pm.003}$ & $0.187^{\pm.002}$ & $0.086^{\pm.001}$ & $0.118^{\pm.003}$ \\
HOI-Diff \cite{hoidiff} &$3.61$s & $0.457^{\pm.003}$ & $0.295^{\pm.003}$ & $0.441^{\pm.005}$ & $0.539^{\pm.006}$ & $0.148^{\pm.003}$ & $0.102^{\pm.000}$ & $0.125^{\pm.002}$\\
HOI-Diff + AIC \cite{hoidiff} & $5.97$s & $0.437^{\pm.004}$ & $0.312^{\pm.002}$ & $0.467^{\pm.003}$ & $0.563^{\pm.006}$ & $0.101^{\pm.001}$ & $\underline{0.081}^{\pm.001}$ & $0.098^{\pm.002}$ \\ \hline
\textbf{Our ChainHOI} & \pmb{$0.61$s} & $\underline{0.095}^{\pm.001}$ & $\underline{0.435}^{\pm.009}$ & $\underline{0.621}^{\pm.011}$ & $\underline{0.717}^{\pm.008}$ & $0.091^{\pm.001}$ & $\underline{0.081}^{\pm.001}$ & $\underline{0.063}^{\pm.000}$\\ 
\textbf{Our ChainHOI + AIC } & \underline{$1.06$s} & $\pmb{0.093}^{\pm.001}$ & $\pmb{0.444}^{\pm.008}$ & $\pmb{0.623}^{\pm.010}$ & $\pmb{0.722}^{\pm.011}$ & $\pmb{0.072}^{\pm.001}$ & $0.099^{\pm.011}$ & $\pmb{0.058}^{\pm.001}$\\ \toprule
\rowcolor{gray!30} \multicolumn{9}{l}{\textit{On the OMOMO dataset}}\\ \hline 
MDM$^{finetuned}$ \cite{mdm} &  & $0.164^{\pm.004}$ & $0.123^{\pm.006}$ & $0.208^{\pm.006}$ & $0.278^{\pm.007}$ & - & - & -\\
MDM$^\star$ \cite{mdm} &  & $0.169^{\pm.005}$ & $0.120^{\pm.004}$ & $0.208^{\pm.006}$ & $0.281^{\pm.009}$ & $0.560^{\pm.003}$ & $0.022^{\pm.006}$ & $0.134^{\pm.001}$\\
PriorMDM$^\star$ \cite{priorMDM}&  & $0.329^{\pm.001}$ & $0.147^{\pm.004}$ & $0.219^{\pm.007}$ & $0.277^{\pm.005}$ & $0.588^{\pm.019}$ & $0.025^{\pm.001}$ & $0.115^{\pm.007}$\\
InterDiff \cite{interdiff}&  & $0.253^{\pm.007}$ & $0.118^{\pm.009}$ & $0.210^{\pm.009}$ & $0.281^{\pm.007}$ & $0.472^{\pm.002}$ & $\pmb{0.015}^{\pm.001}$ & $0.139^{\pm.001}$\\
CHOIS$^\star$ \cite{chois} & & $0.251^{\pm.013}$ & $0.133^{\pm.003}$ & $0.254^{\pm.002}$ & $0.343^{\pm.003}$ & $0.323^{\pm.002}$ & $0.021^{\pm.001}$ & $0.151^{\pm.004}$ \\
HOI-Diff \cite{hoidiff}&  & $0.480^{\pm.001}$ & $0.114^{\pm.002}$ & $0.198^{\pm.003}$ & $0.268^{\pm.002}$ & $0.678^{\pm.005}$ & $0.022^{\pm.002}$ & $0.161^{\pm.001}$\\
HOI-Diff + AIC \cite{hoidiff}&  & $0.245^{\pm.001}$ & $0.140^{\pm.002}$ & $0.253^{\pm.004}$ & $0.340^{\pm.001}$ & $0.301^{\pm.027}$ & $\underline{0.017}^{\pm.001}$ & $0.136^{\pm.004}$\\ \hline
\textbf{Our ChainHOI} &  &   $\underline{0.112}^{\pm.004}$ & $\underline{0.264}^{\pm.005}$ & $\underline{0.431}^{\pm.008}$ & $\underline{0.545}^{\pm.008}$ & $\underline{0.263}^{\pm.002}$ & $0.019^{\pm.001}$ & $\pmb{0.089}^{\pm.009}$\\ 
\textbf{Our ChainHOI + AIC } & & $\pmb{0.098}^{\pm.002}$ & $\pmb{0.266}^{\pm.005}$ & $\pmb{0.434}^{\pm.008}$ & $\pmb{0.549}^{\pm.008}$ & $\pmb{0.120}^{\pm.001}$ & $0.021^{\pm.001}$ & $\underline{0.090}^{\pm.002}$\\ \toprule
\end{tabular}
\vspace{-0.3cm}
\caption{\textbf{Quantitative evaluation of the BEHAVE \cite{behave} and OMOMO \cite{omomo} test sets.} We repeated evaluation 20 times to calculate the average results with a 95\% confidence interval (denoted by ±). The best result is in bold, and the second best is underlined. The Average Inference Time (AIT) is the mean over 100 samples on an RTX 3090. We evaluate the AIT of methods only on the BEHAVE dataset. Affordance-guided Interaction Correction (AIC) \cite{hoidiff} is a post-processing method.}
\label{tab:main}
\end{table*}

We also adopt a loss to constrain the positions of objects. We directly minimize the distance between the predicted object's 6-DoF and the GT 6-DoF:
\begin{equation}
    \mathcal{L}_o = \sum_{i=1}^{L} \left\| \phi_o(O^{pred}_i) - \phi_o (O^{gt}_i) \right\|^2_2,
\end{equation}
where $\phi_o(O^{pred}_i)$ denotes obtaining the predicted object's 6-DoF.

In summary, we use three losses to optimize our ChainHOI model:
\begin{equation}
    \mathcal{L} = \mathcal{L}_{diff} + \lambda_1 \mathcal{L}_h + \lambda_2 \mathcal{L}_o,
\end{equation}
where $\mathcal{L}_{diff}$ is the diffusion loss, and $\lambda_1$ and $\lambda_2$ are hyperparameters.
Refer to the Appendix for more details.

\section{Experiments}

\subsection{Datasets}
We conduct experiments on two publicly available 3D HOI datasets: BEHAVE \cite{behave} and OMOMO \cite{omomo}. The BEHAVE dataset includes 20 objects and 8 subjects, with 1,451 3D HOI sequences, each annotated with three text descriptions \cite{hoidiff}. We use the BEHAVE dataset, which has been preprocessed by HOI-Diff \cite{hoidiff}, and follow the official train-test split. The OMOMO dataset provides 10 hours of HOI sequences with text descriptions, including 15 objects and 17 subjects. We follow the same preprocessing pipeline as BEHAVE, based on HOI-Diff, with a train-test split that ensures distinct objects in the training and testing sets.

\begin{figure*}[ht]
    \centering
    \setlength{\tabcolsep}{0cm}
    \begin{tabular}{cccc}
    \toprule
    \makebox[0.25\textwidth][c]{\textbf{Our ChainHOI}} & \makebox[0.25\textwidth][c]{PriorMDM} & \makebox[0.25\textwidth][c]{InterDiff} & \makebox[0.23\textwidth][c]{HOI-Diff} \\ \hline
    \multicolumn{4}{c}{A person \textbf{sits on a yogaball}.} \\
    \multicolumn{4}{c}{\includegraphics[width=\linewidth]{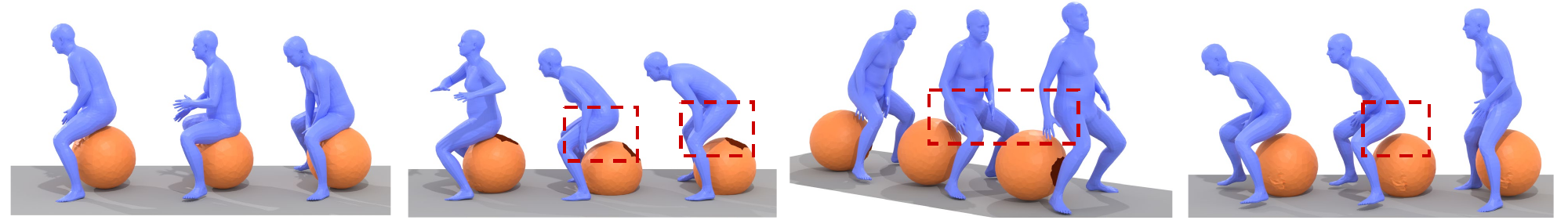}} \vspace{-0.1cm }\\ \hline
    \multicolumn{4}{c}{A person is \textbf{carrying out} upper body exercises employing a \textbf{yogaball}.} \\
    \multicolumn{4}{c}{\includegraphics[width=\linewidth]{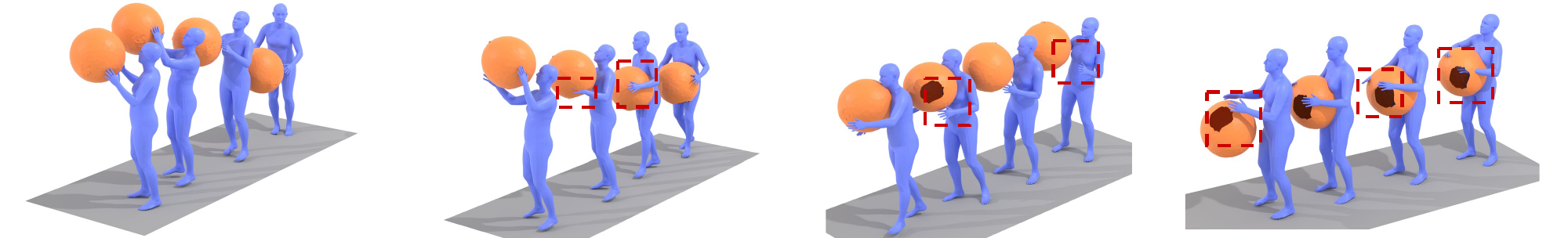}} \vspace{-0.1cm }\\ \hline
    \multicolumn{4}{c}{A person is \textbf{grasping} the \textbf{backpack} with his \textbf{left hand}.} \\
    \multicolumn{4}{c}{\includegraphics[width=\linewidth]{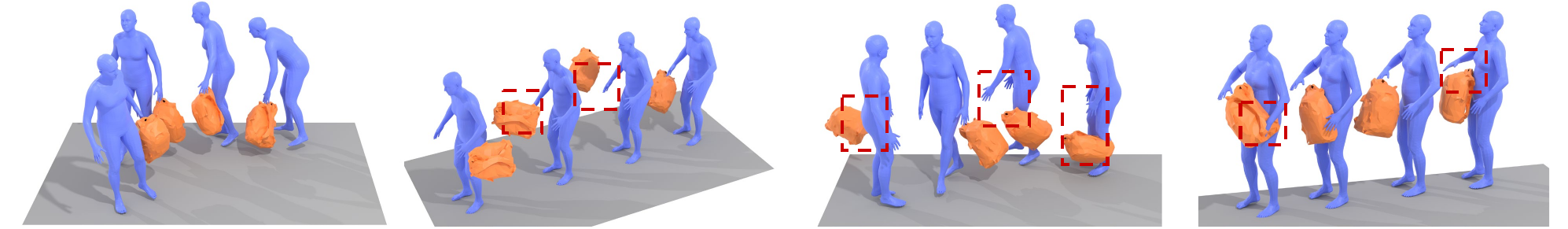}} \vspace{-0.1cm }\\ 
    \toprule
    \end{tabular}
    \vspace{-0.3cm}
    \caption{\textbf{Qualitative comparisons on the BEHAVE dataset.} Only keyframes are shown for clarity. Regions enclosed in red boxes highlight issues, such as mesh clipping or excessively large contact distances. Compared to other methods, our approach generates more realistic and plausible human-object interactions.}
    \label{fig:vis}
\end{figure*}

\subsection{Evaluation Metrics}
\textbf{Evaluating Motion Generation Quality.}
We adopt the metrics from T2M \cite{t2m}: \textit{Frechet Inception Distance} (FID) and \textit{R-Precision}. Both metrics are computed as described in T2M \cite{t2m}. FID measures the similarity between generated and ground truth motions. \textit{R-Precision} assesses the semantic alignment between text descriptions and generated motions. However, computing these metrics requires a pre-trained model to extract features from both motion sequences and text descriptions. Since there is a domain gap between text-driven motion generation and HOI generation, and since HOI-Diff does not provide such a model, we trained a new feature extractor for evaluation, which will be public. Additionally, we use the \textit{Foot Skating Ratio} (FSR) to determine the percentage of frames where foot skid exceeds 2.5 cm during ground contact (i.e., when the foot height is below 5 cm), following HOI-Diff.

\vspace{0.15cm}
\noindent\textbf{Evaluating Interaction Quality.} 
We introduce two metrics: \textit{Penetration Score} (PS) and \textit{Optimal Contact Distance} (OCD). PS measures the proportion of vertices in the human body mesh whose SDF value relative to the object mesh is negative. To compute \textit{Contact Distance}, ground truth (GT) contact labels are required. Unlike prior works \cite{hoidiff, hoianimator, omomo, chois} that use a single GT label, \textit{we argue that multiple motions can align with the same text, making a single GT label insufficient for a generation task}, inspired by \cite{weigrasp, cui2024anyskill}. Therefore, OCD uses ChatGPT 4o to identify all semantically consistent HOI sequences involving the same objects, computes contact distances for each using their contact labels, and selects the smallest distance. 


\subsection{Quantitative Evaluation}

\noindent\textbf{Compared Methods.}
Due to the limited existing methods for text-driven HOI generation (some of which are not open source) and inconsistent data representation and text annotations in the BEHAVE dataset, we reproduce both the available open-source HOI methods and certain text-driven motion generation methods in our HOI representation format for comparison. As our HOI representation can be converted to the format used in HOI-Diff, we directly evaluate the released HOI-Diff checkpoints with our metrics.

We implement the following methods for the text-driven HOI generation task: (1) InterDiff \cite{interdiff}: We extend InterDiff \cite{interdiff}, originally a general motion diffusion model, to be conditioned on text for HOI generation. (2) MDM$^{\text{finetuned}}$: As MDM \cite{mdm} is a text-driven motion method, we directly fine-tune the pretrained MDM model on HOI datasets that focus solely on human motion generation. (3) MDM$^\star$: We concatenate human motion and object 6-DoF along the time dimension and then train MDM from scratch, enabling the HOIs generation. (4) PriorMDM$^\star$ \cite{priorMDM}: Designed for two-person motion generation, PriorMDM$^\star$ replaces one of the two persons with an object to achieve HOI sequence generation. (5) CHOIS$^\star$ \cite{chois}: Because CHOIS uses both text and waypoints, we removed waypoints and adapted its dimensions. Refer to the Appendix for more details.

\vspace{0.15cm}
\noindent\textbf{Quantitative Comparison.}
Experimental results are presented in \cref{tab:main}. We find that while text-driven motion generation methods, \ie, MDM and PriorMDM, perform well in human motion quality, they fall short in interaction quality. In contrast, HOI methods, \ie, InterDiff and HOI-Diff, excel in interaction quality assessment. Our ChainHOI achieves state-of-the-art performance on both datasets, with substantial improvements in FID, R-Precision, OCD, and FSR. Moreover, our ChainHOI is compatible with the Affordance-guided Interaction Correction (AIC) proposed by HOI-Diff. Using AIC to optimize ChainHOI's outputs further enhances its performance. Additionally, the inference time of our ChainHOI is significantly lower than that of existing methods.

\subsection{Qualitative Evaluation}
\cref{fig:vis} presents the qualitative results of different methods. PriorMDM and InterDiff generate unrealistic results, such as excessively large contact distances. Although HOI-Diff produces more reasonable contact distances, it still suffers from significant mesh penetration issues. In contrast, our method outperforms both, generating more realistic and accurate human-object interactions. Please refer to the supplementary materials for complete video visualizations.

\begin{figure}
    \centering
    \includegraphics[width=\linewidth]{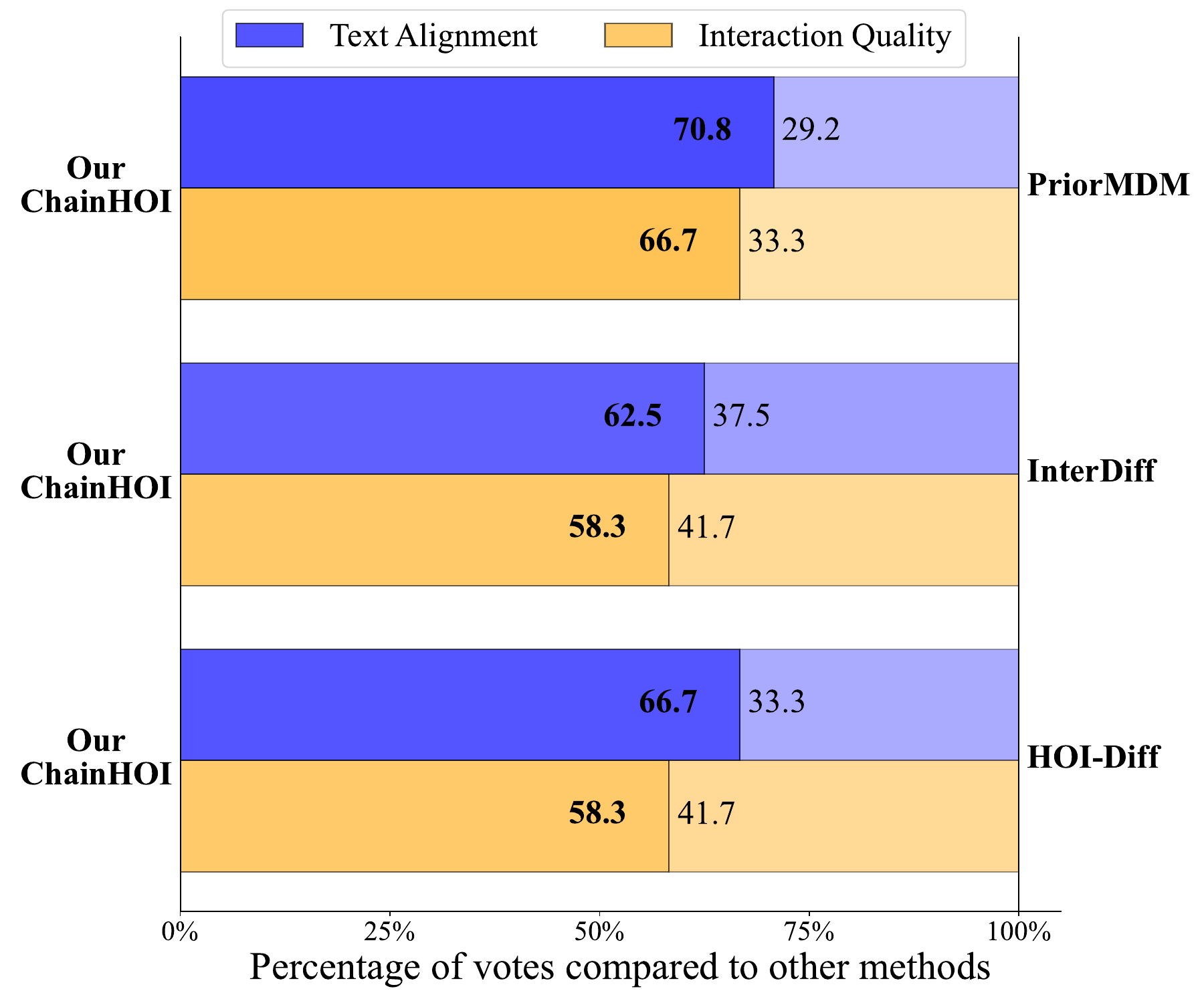}
    \caption{\textbf{User Study.} The color bar and numbers indicate the preference rate of ChainHOI over the compared methods.}
    \label{fig:user study}
\end{figure}

\vspace{0.15cm}
\noindent\textbf{User Study.}
We conduct a user study to evaluate ChainHOI in comparison with existing methods. In side-by-side comparisons of ChainHOI with existing methods, 24 participants were asked to select the one that better aligns with the given text or shows higher interaction quality. \cref{fig:user study} 
 shows that ChainHOI is preferred by participants most of the time. Refer to the Appendix for more details.

\begin{table}[t]
    \centering
    \setlength{\tabcolsep}{0.1cm}
    \resizebox{\linewidth}{!}{
    \begin{tabular}{cccccc}
    \toprule
        & FID$\downarrow$ & R-Top1$\uparrow$ & OCD$\downarrow$ & PS$\downarrow$ \\ \hline
         Transformer & $0.229$ & $0.379$ & $0.251$ & $0.082$ \\
         Transformer$^\star$ & $0.202$ & $0.385$ & $0.338$ & $\pmb{0.073}$\\ \hline
         ST-GCN only & $0.193$ & $0.261$ & $0.235$ & $0.088$\\
         w/o KIM & $\underline{0.142}$ & $0.400$ & $0.116$ & $0.090$\\
         w/o SCM & $0.170$ & $0.380$ & $0.197$ & $\underline{0.077}$\\
         w/o mask in KIM & $0.184$ & $0.428$ & $\underline{0.100}$ & $0.096$\\ \hline
         w/o object geometry & $0.152$ & $\pmb{0.480}$ & $0.278$ & $0.080$\\ \hline
         \rowcolor{gray!30}ChainHOI  & $\pmb{0.095}$ & $\underline{0.435}$ & $\pmb{0.091}$ & $0.081$ \\
    \toprule
    \end{tabular}
    }
    \vspace{-0.3cm}
    \caption{\textbf{Ablation Studies on the BEHAVE datset.} }
    \label{tab:ablation}
\end{table}

\subsection{Ablation Study}
\begin{figure}
    \centering
    \includegraphics[width=\linewidth]{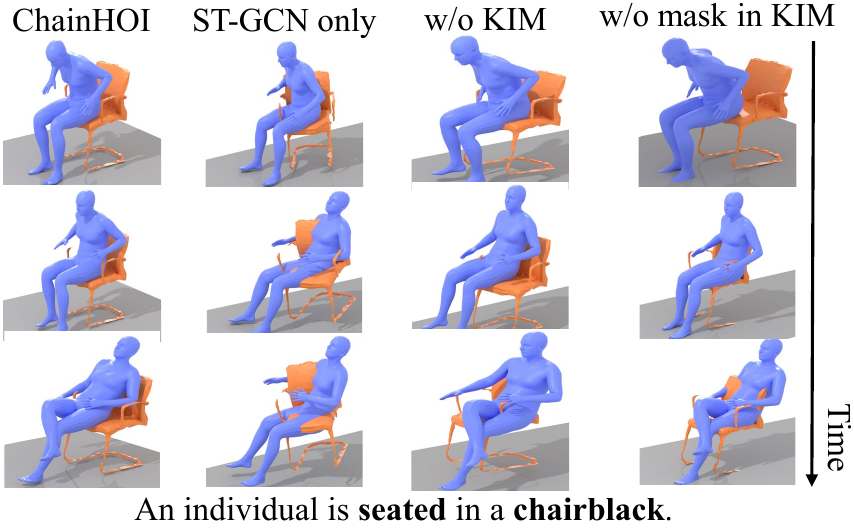}
    \vspace{-0.6cm}
    \caption{\textbf{Visualization Results of Ablation Studies.} Ablating specific modules degrades the quality of the generated results. 
    }
    \label{fig:ablation}
\end{figure}

As shown in \cref{tab:ablation} and \cref{fig:ablation}, we evaluate the impact of different model designs. First, we assess the Transformer model with both implicit and explicit joint-level interaction modeling using two HOI representations: $\mathbf{m} \in \mathbb{R}^{L \times (J \times D_{\text{in}})}$ and $\mathbf{m} \in \mathbb{R}^{L \times J \times D_{\text{in}}}$. The results indicate that even with joint-level representations, using only a Transformer struggles to achieve strong performance.

\vspace{0.15cm}
\noindent\textbf{Effectiveness of Our GST-GCN.}
The model using only ST-GCN performs poorly on R-Top1, with severe a significantly higher OCD. Adding the Semantic-consistent Module (\textit{w/o KIM}) improves R-Top1 from 0.261 to 0.400. Similarly, removing the Semantic-consistent Module alone (\textit{w/o SCM}) causes a significant performance drop, highlighting the importance of long-term information modeling.

\vspace{0.15cm}
\noindent\textbf{Effectiveness of Our KIM.}
There is a significant performance gap in the FID metric between the model without KIM (\textit{w/o KIM}) and our full model, demonstrating KIM's effectiveness. Removing explicit kinetic chains modeling by eliminating the attention mask in KIM still maintains good performance on OCD, but the FID score worsens from 0.095 to 0.184. This underscores the necessity of explicitly modeling kinetic chains, as also shown in \cref{fig:ablation}.

\vspace{0.15cm}
\noindent\textbf{Effectiveness of Object Geometry.}
Unexpectedly, removing object geometry boosts R-Top1 performance but severely degrades OCD. This suggests that without object geometric information, the model struggles to capture detailed interactions, focusing only on human-related aspects.


\section{Conclusion}

We introduced ChainHOI, a method that explicitly models interactions at the joint and kinetic chain levels for text-driven human-object interaction generation. By employing a novel joint graph with the Generative Spatiotemporal Graph Convolution Network (GST-GCN) and the Kinematics-based Interaction Module (KIM), our approach enhances realism and semantic consistency in generated motions. Experiments on BEHAVE and OMOMO datasets demonstrate the effectiveness of ChainHOI, confirming the advantages of explicit interaction modeling.

\section{Acknowledgments}
This work was supported partially by NSFC(92470202, U21A20471), National Key Research and Development Program of China (2023YFA1008503), Guangdong NSF Project (No. 2023B1515040025).

{
    \small
    \bibliographystyle{ieeenat_fullname}
    \bibliography{main}
}

\clearpage

\appendix

\twocolumn[{
\begin{center}
  \Large\bfseries Appendix of ``ChainHOI: Joint-based Kinematic Chain Modeling for \\Human-Object Interaction Generation'' 
  \vspace{1em} 
\end{center}
}] 


\section{Summary}

In the appendix, we first introduce more details of our method and experiments in \cref{sec:details}. Then, we present the full experimental results using additional metrics and analyze these results in \cref{sec:results}. Moreover, we conduct extensive ablation studies to evaluate the impact of each design choice and hyperparameter in \cref{sec:ablation} and \cref{sec:visualization}. In \cref{sec:failure-cases}, we introduce failure cases generated by our method. Finally, we discuss the limitations of our ChainHOI in \cref{sec: limitations}.

\section{More Details of Our Method and Experiments} \label{sec:details}

\subsection{Implementation Details}
$T$ is set to 1000 as the maximum diffusion step, and the variances $\beta_t$ vary from 0.0001 to 0.02. We use DDIM \cite{ddim} with 50 time steps during sampling. The number of blocks $N$ is set to 6. $D_m$ and $D_t$ are set to 64 and 256, respectively. We downsample the number of object points to 16 using PointNet \cite{pointnet}. Our ChainHOI is optimized using AdamW \cite{adamw} on two RTX 3090 Ti GPUs in parallel with a learning rate of 1e-4 and a batch size of 32. The model is trained for 200 epochs. During testing, the guidance scale is set to 2. $\lambda_1$ and $\lambda_2$ are set to 2 and 1, respectively.

\subsection{Details of Our Loss}
We note that non-watertight objects do not affect the computation of $\mathcal{G}()$ during training because $\mathcal{G}()$ computes the square of the minimum absolute distance from the joint to all triangles.

\noindent\textbf{Why is the $\mathcal{G}()$ calculated from human joints to the ground truth object rather than the generated object?} 
Because the object consists of a large number of triangular facets, using the generated object information results in a 3.6-fold increase in GPU memory usage (1.8 GB vs. 6.6 GB when the batch size is 1).  We evaluate the performance using generated objects, reducing the batch size to one-fourth of the original due to GPU memory constraints. \cref{tab:loss_comp} shows that performance was inferior compared to using GT objects.

\begin{table}[h]
    \centering
   \begin{tabular}{c|cccc}
    \bottomrule[1pt]
     & FID$\downarrow$ & R-Top1$\uparrow$ & OCD$\downarrow$ & PS$\downarrow$ \\ \hline
    
    
     {gen obj.} &  {$0.098$} &  {$\pmb{0.437}$} &  {$\pmb{0.089}$} &  {$\pmb{0.081}$} \\ 
    
    \rowcolor{gray!30}  {GT obj.} &  {$\pmb{0.095}$} &  {${0.435}$} &  {$0.091$} &  {$\pmb{0.081}$} \\ \toprule[1pt]
    \end{tabular}
    \caption{Comparisons of using the generated object or the ground truth object to calculate the distance.}
    \label{tab:loss_comp}
\end{table}

\subsection{Details of Our OCD}

LLM-assisted label generation and evaluation have been extensively utilized in recent studies \cite{cui2024anyskill, li2024semgrasp, dubois2023alpacafarm, weigrasp}. Specifically, we first filter grouping candidates by object category and contacting body parts (avg. 5.1 candidates on average in each group). Next, we utilize ChatGPT-4o to determine which instructions are semantically identical by evaluating action intent and the specific human body part involved in the contact. Consequently, this task is relatively straightforward for ChatGPT-4o. \Cref{tab:prompt_group} shows the prompt used to group semantically identical HOIs.

We also assess the quality of the labeling process through two methods: manual evaluation and LLM-assisted evaluation. First, we review a 10\% sample of the labels, achieving an accuracy rate of 94\%. Second, following the approach in \cite{li2024semgrasp}, we employ ChatGPT-4o to evaluate all labels to ensure consistency within each instruction group, resulting in a mean consistency score of 0.906 (on a scale from 0 to 1). \Cref{tab:prompt_check} shows the prompt used to evaluate group labels to ensure consistency within each instruction group.

\subsection{Details of Our User Study}
We initiate the evaluation by randomly selecting 20 test prompts from the BEHAVE dataset. Subsequently, for each of these 20 prompts, we instruct each method to generate 5 samples. This process yields a corpus of 100 samples, which are then used in a pairwise user study. The evaluation score is calculated as the ratio of votes received to the total votes cast.

\subsection{Details of Our Evaluator} \label{sec:evaluator}
As mentioned in our main manuscript, we adopt the metrics from T2M \cite{t2m} to evaluate motion generation quality. However, computing these metrics requires a pre-trained model to extract features from both motion sequences and text descriptions. Since there is a domain gap between text-driven motion generation and HOI generation, and because HOI-Diff does not provide such a model, we trained a new feature extractor for evaluation.

\begin{figure}
    \centering
    \includegraphics[width=\linewidth]{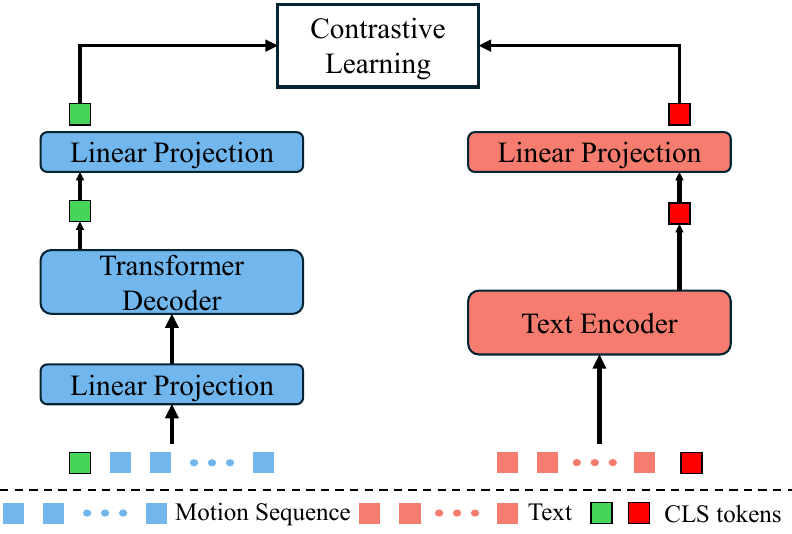}
    \caption{\textbf{Overview of Our Evaluator.} Inspired by CLIP \cite{clip}, our evaluator incorporates a motion branch and a text branch. The motion branch takes motion sequences and a CLS token as inputs. The text branch takes texts and a CLS tokens as inputs. Then the output CLS tokens of two branches are passed into a linear projection and then are used to calculate the contrastive learning loss.}
    \label{fig:evaluator}
\end{figure}

Inspired by CLIP \cite{clip}, we design and train our evaluator using a contrastive learning method. As shown in Figure \ref{fig:evaluator}, our evaluator consists of a motion branch and a text branch. The motion branch takes motion sequences and a CLS token as inputs, while the text branch takes texts and a CLS token as inputs. The output CLS tokens from both branches are then passed through a linear projection and used to calculate the contrastive learning loss.

\vspace{0.2cm}
\noindent\textbf{Implementation Details.}
Following previous works \cite{chois, hoidiff, cghoi, hoianimator}, our evaluator is used to assess the quality of human motion only (excluding object motion). During training, human motions are represented using the HumanML3D representation \cite{t2m} $\bar{\mathbf{m}} \in \mathbf{R}^{L \times D}$, where $D=263$. The motion branch consists of 8 Transformer Decoders \cite{transformer}, while the text branch uses the pretrained RoBERTa \cite{roberta}. The dimensionality of the motion branch is 384. The output dimensions of the linear projections for both branches are 512. Our evaluator is optimized using the Adam optimizer \cite{adam} with a learning rate of $1 \times 10^{-4}$ on an RTX 3090 Ti GPU. The batch size and number of training epochs are set to 64 and 200, respectively. Specifically, our evaluator is trained in two stages. In the first 8 epochs, the text branch, except for the linear projection, is fixed, and only the motion branch is trained. Then, all branches are trained together.

\begin{table*}[h]
\centering
\setlength\tabcolsep{1mm}
\resizebox{\linewidth}{!}{
\begin{tabular}{lcccccccccc}
\bottomrule
 \multirow{2}{*}{Methods} & \multirow{2}{*}{FID$\downarrow$}  & \multicolumn{3}{c}{R-Precision$\uparrow$} & \multirow{2}{*}{MM. Dist.$\downarrow$} & \multirow{2}{*}{Div.$\uparrow$} & \multirow{2}{*}{OCD $\downarrow$} & \multirow{2}{*}{PS $\downarrow$} & \multirow{2}{*}{FSR $\downarrow$} & \multirow{2}{*}{CD $\downarrow$}\\ \cline{3-5}
   & & Top1 & Top2 & Top3 & & \\ \toprule
\rowcolor{gray!30} \multicolumn{11}{l}{\textit{On the BEHAVE dataset}}\\ \hline 
Real motion. & $0.001^{\pm.000}$ & $0.287^{\pm.011}$ & $0.443^{\pm.013}$ & $0.544^{\pm.011}$ & $1.055^{\pm.002}$ & $1.346^{\pm.011}$ & - & - & - & -\\ \hline
MDM$^{finetuned}$ \cite{mdm} & $0.246^{\pm.006}$ & $0.223^{\pm.011}$ & $0.378^{\pm.105}$ & $0.488^{\pm.015}$ & $1.118^{\pm.006}$ & $\pmb{1.350}^{\pm.008}$ & - & - & - & - \\
MDM$^\star$ \cite{mdm} & $0.257^{\pm.004}$ & $0.220^{\pm.007}$ & $0.355^{\pm.001}$ & $0.451^{\pm.001}$ & $1.071^{\pm.003}$ & $1.307^{\pm.006}$ & $0.458^{\pm.013}$ & $0.095^{\pm.007}$ & $0.098^{\pm.002}$ & $0.481^{\pm.014}$\\
PriorMDM$^\star$ \cite{priorMDM}& $0.328^{\pm.018}$ & $0.243^{\pm.009}$ & $0.329^{\pm.009}$ & $0.385^{\pm.013}$ & $1.203^{\pm.012}$ & $1.142^{\pm.017}$ & $0.215^{\pm.012}$ & $0.116^{\pm.001}$ & $0.066^{\pm.004}$ & $0.232^{\pm.010}$\\
InerDiff \cite{interdiff} & $0.170^{\pm.002}$ & $0.310^{\pm.003}$ & $0.480^{\pm.005}$ & $0.599^{\pm.001}$ & $1.045^{\pm.002}$ & $1.325^{\pm.006}$ & $0.191^{\pm.027}$ & $\pmb{0.078}^{\pm.000}$ & $0.069^{\pm.002}$ & $0.206^{\pm.003}$\\
CHOIS$^\star$ \cite{chois} & $0.157^{\pm.001}$ & $0.301^{\pm.002}$ & $0.488^{\pm.003}$ & $0.606^{\pm.003}$ & $1.090^{\pm.002}$ & $1.265^{\pm.013}$ & $0.187^{\pm.002}$ & $0.086^{\pm.001}$ & $0.118^{\pm.003}$ & $0.202^{\pm.003}$\\
HOI-Diff \cite{hoidiff}& $0.457^{\pm.003}$ & $0.295^{\pm.003}$ & $0.441^{\pm.005}$ & $0.539^{\pm.006}$ & $1.119^{\pm.009}$ & $1.204^{\pm.017}$ &  $0.148^{\pm.003}$ & $0.102^{\pm.000}$ & $0.125^{\pm.002}$ & $0.214^{\pm.003}$\\
HOI-Diff + AIC \cite{hoidiff}& $0.437^{\pm.004}$ & $0.312^{\pm.002}$ & $0.467^{\pm.003}$ & $0.563^{\pm.006}$ & $1.107^{\pm.003}$ & $1.235^{\pm.020}$ & $0.101^{\pm.001}$ & $\underline{0.081}^{\pm.001}$ & $0.098^{\pm.002}$ & $0.117^{\pm.003}$\\ \hline
\textbf{Our ChainHOI} & $\underline{0.095}^{\pm.001}$ & $\underline{0.435}^{\pm.009}$ & $\underline{0.621}^{\pm.011}$ & $\underline{0.717}^{\pm.008}$ & $\underline{0.967}^{\pm.001}$ & $\underline{1.337}^{\pm.015}$ &
$0.091^{\pm.001}$ & $\underline{0.081}^{\pm.001}$ & $\underline{0.063}^{\pm.000}$ & $\underline{0.096}^{\pm.001}$\\ 
\textbf{Our ChainHOI + AIC } & $\pmb{0.093}^{\pm.001}$ & $\pmb{0.444}^{\pm.008}$ & $\pmb{0.623}^{\pm.010}$ & $\pmb{0.722}^{\pm.011}$ & $\pmb{0.964}^{\pm.004}$ & $\underline{1.337}^{\pm.010}$ & $\pmb{0.072}^{\pm.001}$ & $0.099^{\pm.011}$ & $\pmb{0.058}^{\pm.001}$ & $\pmb{0.078}^{\pm.001}$\\ \toprule
\rowcolor{gray!30} \multicolumn{11}{l}{\textit{On the OMOMO dataset}}\\ \hline 
Real motion. & $0.001^{\pm.001}$ & $0.247^{\pm.006}$ & $0.398^{\pm.004}$ & $0.504^{\pm.005}$ & $1.050^{\pm.001}$ & $1.356^{\pm.013}$ & - & - & - & -\\ \hline
MDM$^{finetuned}$ \cite{mdm}  & $0.164^{\pm.004}$ & $0.123^{\pm.006}$ & $0.208^{\pm.006}$ & $0.278^{\pm.007}$ & $1.228^{\pm.004}$ & $1.333^{\pm.002}$ & - & - & - & -\\
MDM$^\star$ \cite{mdm}  & $0.169^{\pm.005}$ & $0.120^{\pm.004}$ & $0.208^{\pm.006}$ & $0.281^{\pm.009}$ & $1.191^{\pm.004}$ & $1.319^{\pm.001}$ & $0.560^{\pm.003}$ & $0.022^{\pm.006}$ & $0.134^{\pm.001}$ & $0.686^{\pm.002}$\\
PriorMDM$^\star$ \cite{priorMDM}& $0.329^{\pm.001}$ & $0.147^{\pm.004}$ & $0.219^{\pm.007}$ & $0.277^{\pm.005}$ & $1.200^{\pm.005}$ & $1.181^{\pm.003}$ & $0.588^{\pm.019}$ & $0.025^{\pm.001}$ & $0.115^{\pm.007}$ & $0.755^{\pm.022}$\\
InerDiff \cite{interdiff}& $0.253^{\pm.007}$ & $0.118^{\pm.009}$ & $0.210^{\pm.009}$ & $0.281^{\pm.007}$ & $\underline{1.167}^{\pm.001}$ & $1.227^{\pm.003}$ & $0.472^{\pm.002}$ & $\pmb{0.015}^{\pm.001}$ & $0.139^{\pm.001}$ & $0.585^{\pm.003}$\\
CHOIS$^\star$ \cite{chois} & $0.251^{\pm.013}$ & $0.133^{\pm.003}$ & $0.254^{\pm.002}$ & $0.343^{\pm.003}$ & $1.193^{\pm.003}$ & $1.334^{\pm.014}$ & $0.323^{\pm.002}$ & $0.021^{\pm.001}$ & $0.151^{\pm.004}$ & $0.433^{\pm.001}$\\
HOI-Diff \cite{hoidiff}& $0.480^{\pm.001}$ & $0.114^{\pm.002}$ & $0.198^{\pm.003}$ & $0.268^{\pm.002}$ & $1.221^{\pm.008}$ & $1.124^{\pm.020}$ & $0.678^{\pm.005}$ & $0.022^{\pm.002}$ & $0.161^{\pm.001}$ & $0.763^{\pm.014}$\\
HOI-Diff + AIC \cite{hoidiff}& $0.245^{\pm.001}$ & $0.140^{\pm.002}$ & $0.253^{\pm.004}$ & $0.340^{\pm.001}$ & $1.183^{\pm.005}$ & $1.303^{\pm.014}$ & $0.301^{\pm.027}$ & $\underline{0.017}^{\pm.001}$ & $0.136^{\pm.004}$ & $0.331^{\pm.015}$\\ \hline
\textbf{Our ChainHOI}&   $\underline{0.112}^{\pm.004}$ & $\underline{0.264}^{\pm.005}$ & $\underline{0.431}^{\pm.008}$ & $\underline{0.545}^{\pm.008}$ & $\pmb{1.023}^{\pm.007}$ & $\pmb{1.350}^{\pm.002}$ & $\underline{0.263}^{\pm.002}$ & $0.019^{\pm.001}$ & $\pmb{0.089}^{\pm.009}$ & $\underline{0.283}^{\pm.009}$\\ 
\textbf{Our ChainHOI + AIC } & $\pmb{0.098}^{\pm.002}$ & $\pmb{0.266}^{\pm.005}$ & $\pmb{0.434}^{\pm.008}$ & $\pmb{0.549}^{\pm.008}$ & $\pmb{1.023}^{\pm.007}$ & $\underline{1.348}^{\pm.018}$ & $\pmb{0.120}^{\pm.001}$ & $0.021^{\pm.001}$ & $\underline{0.090}^{\pm.002}$ & $\pmb{0.136}^{\pm.009}$\\ \toprule
\end{tabular}
}
\caption{\textbf{Quantitative evaluation of the BEHAVE \cite{behave} and OMOMO \cite{omomo} test sets.} We repeated evaluation 20 times to calculate the average results with a 95\% confidence interval (denoted by ±). The best result is in bold, and the second best is underlined.  Affordance-guided Interaction Correction (AIC) \cite{hoidiff} is a post-processing method.}
\label{tab:main2}
\end{table*}

\subsection{Details of Compared Baselines} \label{sec:baselines}

Apart from the methods compared in our main manuscript, we have included a modified version of CHOIS \cite{chois} for comparison, as it recently released its source code. Due to space constraints, we provide a brief description of these modifications in the main manuscript. Below, we provide a detailed description of the modified methods:

\begin{itemize}
    \item CHOIS$^\star$ \cite{chois}: Since CHOIS aims to generate HOI sequences conditioned on both text descriptions and object waypoints, we removed the object waypoints from the CHOIS model and modified the input and output dimensions. Note that our HOI representation is easily compatible with the one used in CHOIS.
    \item InterDiff \cite{interdiff}: InterDiff is designed for HOI prediction conditioned on past HOI sequences. To adapt it to text-driven HOI generation, we replaced the past HOI sequences with text descriptions. Specifically, we modified the feature dimensions and utilized the CLIP text encoder to extract text features.
    \item MDM$^{finetuned}$ \cite{mdm}: As MDM is a text-driven motion generation method, we directly fine-tuned the MDM model pretrained on the HumanML3D dataset \cite{t2m} to generate human motion only. Note that MDM$^{finetuned}$ does not generate object motion, and metrics for evaluating interaction quality are not included.
    \item MDM$^\star$ \cite{mdm}: To adapt MDM to the text-driven HOI generation task, we concatenated object motion (6-DoF) and human motion as the model's inputs and outputs. This modification allows MDM$^\star$ to generate HOI sequences.
    \item PriorMDM$^\star$ \cite{priorMDM}: PriorMDM introduces a ComMDM block for two-person motion generation. To adapt it to the text-driven HOI generation task, we replaced one of the two persons with the object and modified the input and output dimensions accordingly.
\end{itemize}

\section{Experiments Using More Metrics} \label{sec:results}

In this section, we present comprehensive results evaluated using additional metrics to demonstrate the effectiveness of our ChainHOI. In addition to the metrics introduced in our manuscript, we utilize the following metrics to assess generation quality:
\begin{itemize}
    \item \textit{MultiModal Distance} (MM. Dist.): MM. Dist. calculates the average distance between the motion features of each generated motion and the text features of their corresponding descriptions in the test set. Note that the features for both motion and text are extracted by our evaluator.
    \item \textit{Diversity} (Div.): Div. measures the variance in the generated motions. We randomly sample two equal-sized subsets from all motions and then compute the average distance between these subsets.
    \item \textit{Contact Distance} (CD): We also report the original contact distance used in previous HOI generation methods. In contrast to our optimal contact distance, the original contact distance uses contact labels from a single ground-truth label.
\end{itemize}

The complete experimental results are presented in \cref{tab:main2}. These findings demonstrate that our ChainHOI performs well on the newly introduced metrics, namely MM Dist., Div., and CD. For the metrics OCD and CD, different models exhibit similar trends. The gap between OCD and CD indicates that using a single ground-truth contact label to calculate contact distance in a generative model is inappropriate. In contrast, our OCD provides a more accurate evaluation of contact distance. On the other hand, we note that it is not surprising that some models' R-Precisions outperform those of real motions, as such phenomena have been reported in many works, such as \cite{hoianimator, momask, bamm, mmm}.

\section{More Ablation Studies} \label{sec:ablation}

In this section, we conduct extensive ablation studies to evaluate the effectiveness of each component and design choice.

\begin{table}[t]
    \centering
    \begin{tabular}{cccccc}
    \toprule
        & FID$\downarrow$ & R-Top1$\uparrow$ & OCD$\downarrow$ & PS$\downarrow$ \\ \hline
         Discrete Graph & $0.138$ & $\pmb{0.457}$ & $0.121$ & $0.084$ \\
         Complete Graph & $0.154$ & $0.443$ & $0.106$ & $0.086$ \\\hline
         Our HOI Graph  & $\pmb{0.095}$ & $0.435$ & $\pmb{0.091}$ & $\pmb{0.081}$ \\
    \toprule
    \end{tabular}
    \caption{\textbf{Evaluations of different HOI graph designs on the BEHAVE dataset.}}
    \label{tab:joint graph}
\end{table}

\subsection{Impact of the Design of HOI Graph}

To evaluate the effectiveness of our HOI graph, we compare our HOI graph with the following designs:
\begin{itemize}
    \item \textbf{Discrete Graph}: The HOI graph is a discrete graph with no edge connections between any two joints.
    \item \textbf{Complete Graph}: The HOI graph is a complete graph with edges between every pair of joints.
\end{itemize}

The experimental results are shown in \cref{tab:joint graph}. The results indicate that, compared to the discrete graph and complete graph, our HOI graph achieves the best performance across all metrics except for R-Top1. Although the discrete graph and complete graph perform better on R-Top1, the FID of both designs is significantly lower than that of our HOI graph.

\begin{table}[t]
    \centering
    \begin{tabular}{cccccc}
    \toprule
        & FID$\downarrow$ & R-Top1$\uparrow$ & OCD$\downarrow$ & PS$\downarrow$ \\ \hline
         Design A & $0.179$ & $\pmb{0.450}$ & $0.113$ & $0.090$ \\
         Design B & $\pmb{0.095}$ & $0.428$ & $0.103$ & $0.084$ \\\hline
         Our Kinetic Chain  & $\pmb{0.095}$ & $0.435$ & $\pmb{0.091}$ & $\pmb{0.081}$ \\
    \toprule
    \end{tabular}
    \caption{\textbf{Evaluations of different kinetic chain designs on the BEHAVE dataset.}}
    \label{tab:chain graph}
\end{table}
\subsection{Impact of the Design of Kinetic Chains}

To evaluate the effectiveness of our kinetic chain design, we propose two alternative configurations for both the internal kinetic chains and the human-object chain:
\begin{itemize}
    \item \textbf{Design A}: This design modifies the internal kinetic chains by reducing the number from five to two. The two remaining kinetic chains represent the upper and lower body, respectively. Note that the human-object chain remains unchanged in this design.
    \item \textbf{Design B}: This design alters the human-object chain by replacing it with a fully connected graph. Specifically, every human joint is connected to the object node, rather than only the potential interaction joints. The internal kinetic chains remain the same as in our original design.
\end{itemize}

As shown in \cref{tab:chain graph}, Design A achieves higher performance on R-Top1. However, the FID, OCD, and PS metrics all decrease significantly compared to our original design. In contrast, Design B, which connects the object node to all human joints, outperforms Design A in both FID and interaction-related metrics. Overall, the design used in our ChainHOI achieves better FID, OCD, and PS while maintaining good R-Top1 performance.

\begin{table}[t]
    \centering
    \begin{tabular}{ccccccc}
    \toprule
        $\lambda_1$ & $\lambda_2$ & FID$\downarrow$ & R-Top1$\uparrow$ & OCD$\downarrow$ & PS$\downarrow$ \\ \hline
         0 & 1 & 0.126 & 0.449 & 0.094 & 0.085\\
         0.5 & 1 & 0.130 & 0.443 & 0.108 & 0.084\\
         1 & 1 & 0.142 & \pmb{0.466} & 0.095 & 0.084\\
         1.5 & 1 & 0.096 & 0.447 & \pmb{0.089} & 0.083\\
         \rowcolor{gray!30} 2 & 1 & $\pmb{0.095}$ & $0.435$ & $0.091$ & $\pmb{0.081}$\\
         2.5 & 1 & 0.175 & 0.414 & \pmb{0.089} & 0.082\\ \hline
         
         2 & 0 & 0.126 & \pmb{0.482} & 0.092 & 0.087\\
         2 & 0.5 & 0.109 & 0.436 & 0.080 & \pmb{0.081}\\ 
         \rowcolor{gray!30}2 & 1 & $\pmb{0.095}$ & $0.435$ & $0.091$ & $\pmb{0.081}$\\
         2 & 1.5 & 0.142 & 0.432 & \pmb{0.078} & 0.089\\
    \toprule
    \end{tabular}
    \caption{\textbf{Impact of the training loss.} The gray line represents the configuration used in our ChainHOI model.}
    \label{tab:loss}
\end{table}
\subsection{Impact of the Training Loss}
As mentioned in Section 3.5 of our main manuscript, we propose two loss functions to improve the quality of human-object interactions. To analyze the impact of our proposed losses, we evaluate the performance by varying the weight of each loss.

The evaluation results are shown in \cref{tab:loss}. As illustrated in the upper part of \cref{tab:loss}, constraining the distance between the predicted human joints and the ground-truth object mesh significantly improves FID and PS, and slightly improves the performance on OCD. In the lower part of \cref{tab:loss}, explicitly constraining the object's 6-DoF significantly enhances the performance on OCD and PS.

\begin{table}[t]
    \centering
    \begin{tabular}{cccccc}
    \toprule
        & FID$\downarrow$ & R-Top1$\uparrow$ & OCD$\downarrow$ & PS$\downarrow$ \\ \hline
         Shared Decoder & $0.174$ & $\pmb{0.447}$ & $\pmb{0.091}$ & $0.085$\\
         Independent Decoders  & $\pmb{0.095}$ & $0.435$ & $\pmb{0.091}$ & $\pmb{0.081}$ \\
    \toprule
    \end{tabular}
    \caption{\textbf{Evaluations of the design of Semantic-consistent Module and Context-aware Decoder on the BEHAVE dataset.}}
    \label{tab:decoder}
\end{table}
\subsection{Impact of the Design of Semantic-consistent Module and Context-aware Decoder}

As shown in Figure 4 of our main manuscript, both the Semantic-consistent Module and the Context-aware Decoder adopt two Transformer decoders to separately obtain information from object geometry and text (denoted as Independent Decoders). To demonstrate the necessity of using two different Transformer decoders, we evaluate the performance when using a single decoder to obtain information from both object geometry and text simultaneously (denoted as Shared Decoder). Specifically, the object geometry tokens and text tokens are concatenated and then input into a Transformer decoder. The experimental results are shown in \cref{tab:decoder}. When using the Shared Decoder, the FID drops significantly, demonstrating the necessity of using Independent Decoders.

\begin{table}[t]
    \centering
    \begin{tabular}{cccccc}
    \toprule
        \#Points & FID$\downarrow$ & R-Top1$\uparrow$ & OCD$\downarrow$ & PS$\downarrow$ \\ \hline
         8 & $0.159$ & $0.364$ & $\pmb{0.083}$ & $0.091$ \\
         \rowcolor{gray!30} 16 & $\pmb{0.095}$ & $0.435$ & $0.091$ & $\pmb{0.081}$\\
         32  & $0.109$ & $\pmb{0.446}$ & $0.095$ & $0.087$ \\
         64 & $0.124$ & $0.424$ & $0.094$ & $\pmb{0.081}$\\
    \toprule
    \end{tabular}
    \caption{\textbf{Effect of the number of points sampled by PointNet.} The gray line represents the configuration used in our ChainHOI.}
    \label{tab:points}
\end{table}

\subsection{Impact of PointNet}
Our ChainHOI adopts PointNet \cite{pointnet} to extract features from object geometry. To evaluate the impact of the number of points sampled by PointNet, we conducted experiments with varying point counts. The experimental results are shown in \cref{tab:points}. The results indicate that using 8 points results in the worst performance on FID, R-Top1, and PS, while performing well on OCD. Furthermore, we find that increasing the number of points does not lead to higher performance. Therefore, we use 16 points in our ChainHOI.

\begin{table}[t]
    \centering
    \begin{tabular}{cccccc}
    \toprule
        \makecell{Inference \\Steps} & AIT & FID$\downarrow$ & R-Top1$\uparrow$ & OCD$\downarrow$ & PS$\downarrow$ \\ \hline
         20 & \textbf{0.28s} & $0.101$ & $0.434$ & $0.093$ & $0.085$ \\
         \rowcolor{gray!30} 50 & 0.61s& $0.095$ & $0.435$ & $0.091$ & $0.081$\\
         100  & 1.41s & $\pmb{0.093}$ & $0.436$ & $0.090$ & $0.081$ \\
         200 & 2.92s & $\pmb{0.093}$ & $\pmb{0.438}$ & $\pmb{0.089}$ & $\pmb{0.080}$\\
    \toprule
    \end{tabular}
    \caption{\textbf{Impact of the number of inference steps.} The gray line represents the configuration used in our ChainHOI model. The Average Inference Time (AIT) is the mean over 100 samples on an RTX 3090Ti.}
    \label{tab:steps}
\end{table}

\subsection{Impact of Inference Steps}
We also evaluate the impact of the number of inference steps. Note that we use DDIM \cite{ddim} to generate HOI sequences during inference. The experimental results are shown in \cref{tab:steps}. As the number of sampling steps increases, the model's performance also improves. However, considering the inference time cost, we set the number of inference steps to 50 to balance generation quality and inference efficiency.

\begin{table}[t]
    \centering
    \begin{tabular}{cccccc}
    \toprule
        \makecell{Guidance \\Scale}  & FID$\downarrow$ & R-Top1$\uparrow$ & OCD$\downarrow$ & PS$\downarrow$ \\ \hline
         1 & $0.102$ & $0.344$ & $0.100$ & $0.086$ \\
         \rowcolor{gray!30} 2 & $0.095$ & $0.435$ & $\pmb{0.091}$ & $\pmb{0.081}$\\
         3  & $0.095$ & $0.460$ & $0.094$ & $0.083$ \\
         4 & $\pmb{0.094}$ & $0.482$ & $0.102$ & $0.085$\\
         5 & $\pmb{0.094}$ & $\pmb{0.498}$ & $0.114$ & $0.084$\\
    \toprule
    \end{tabular}
    \caption{\textbf{Impact of the guidance scale.} The gray line represents the configuration used in our ChainHOI model.}
    \label{tab:guidance}
\end{table}

\subsection{Impact of Guidance Scale}
We conduct experiments to evaluate the impact of the guidance scale during generation. We adopt the classifier-free method \cite{ho2022classifier} to achieve conditional generation. The evaluation results are presented in \cref{tab:guidance}. When the guidance scale is set to 1, the performance on FID, OCD, and PS is satisfactory, likely because the model utilizes the input object geometry to guide HOI generation. However, without text guidance, the generated HOIs may not correspond to the provided text, leading to a lower R-Top1 score. Conversely, as the guidance scale increases, the performance on FID and R-Top1 improves, while the quality of human-object interactions declines, since the quality of human-object interactions does not depend on text guidance.

\begin{table}[t]
    \centering
    \begin{tabular}{cccccc}
    \toprule
        \#Blocks  & FID$\downarrow$ & R-Top1$\uparrow$ & OCD$\downarrow$ & PS$\downarrow$ \\ \hline
         2 & 0.207 & 0.248 & 0.171 & 0.091\\
         4 & 0.141 & 0.342 & 0.104 & 0.088\\
         \rowcolor{gray!30} 6 & $\pmb{0.095}$ & $\pmb{0.435}$ & $0.091$ & $\pmb{0.081}$\\
         8 & 0.135 & 0.423 & \pmb{0.088} & 0.083\\
    \toprule
    \end{tabular}
    \caption{\textbf{Impact of the number of blocks.} The gray line represents the configuration used in our ChainHOI model.}
    \label{tab:block}
\end{table}

\subsection{Impact of the Number of Blocks}
Furthermore, we evaluate the impact of the number of blocks in our ChainHOI model. As shown in \cref{tab:block}, both increasing and decreasing the number of blocks lead to a performance drop. Therefore, the ChainHOI model using six blocks is our final model.

\subsection{Generalization Performance Evaluation}
To evaluate ChainHOI's generalization performance on unseen objects, we tested our model, pre-trained on the OMOMO dataset, on the 3D-FUTURE dataset \cite{fu20213d} using the protocol established by CHOIS \cite{chois}. Results in \Cref{tab:gen} reveal that although a performance drop is observed for all methods, our approach maintains superior performance compared to others.

\begin{table}[t]
    \centering
    \begin{tabular}{cccccc}
    \toprule
          & FID$\downarrow$ & R-Top1$\uparrow$ & OCD$\downarrow$ & PS$\downarrow$ \\ \hline
       {HOI-Diff} &  {$0.514$} &  {$0.097$} &  {$0.281$} &  {$0.023$} \\
     {CHOIS$^\star$} &  {$0.368$} &  {$0.107$} &  {$0.269$} &  {$0.026$} \\ 
    \rowcolor{gray!30}  {Our} &  {$\pmb{0.154}$} &  {$\pmb{0.186}$} &  {$\pmb{0.238}$} &  {$\pmb{0.022}$} \\ \toprule[1pt]
    \end{tabular}
    \caption{\textbf{Generalization Performance Evaluation.} All methods are trained on the OMOMO dataset and evaluated on the 3D-FUTURE dataset. }
    \label{tab:gen}
\end{table}

\begin{figure*}[ht]
    \centering
    \includegraphics[width=\linewidth]{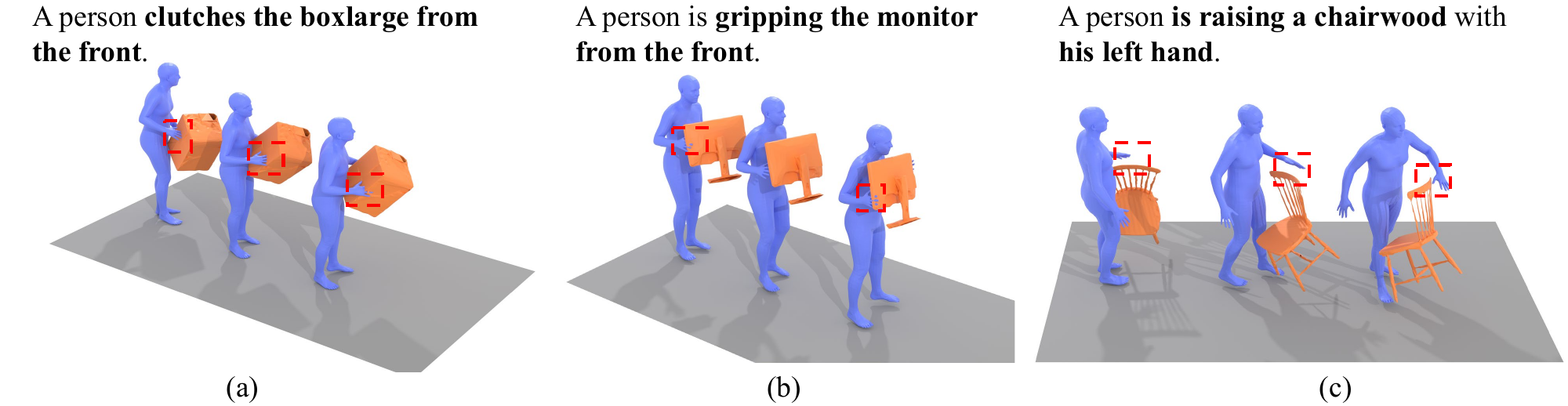}
    \caption{\textbf{Visualization of Failure Cases.} We present two typical failure cases encountered by our method.
    \textit{Issues with Human-Object Clipping}: Subfigures (a) and (b) illustrate problems related to clipping between the fingers and objects. Since the input data is represented using the SMPL model \cite{SMPL}, which does not include finger joint information, our ChainHOI is unable to accurately model the fingers, resulting in clipping between the hands and objects.
    \textit{Large Contact Distances}: Subfigure (c) demonstrates that the contact distance may be excessively large when interacting with certain complex objects. For instance, with objects such as chairs, our model struggles to learn the correct contact points and appropriate contact distances.}
    \label{fig:failure-cases}
\end{figure*}

\begin{figure*}[t]
    \centering
    \includegraphics[width=\linewidth]{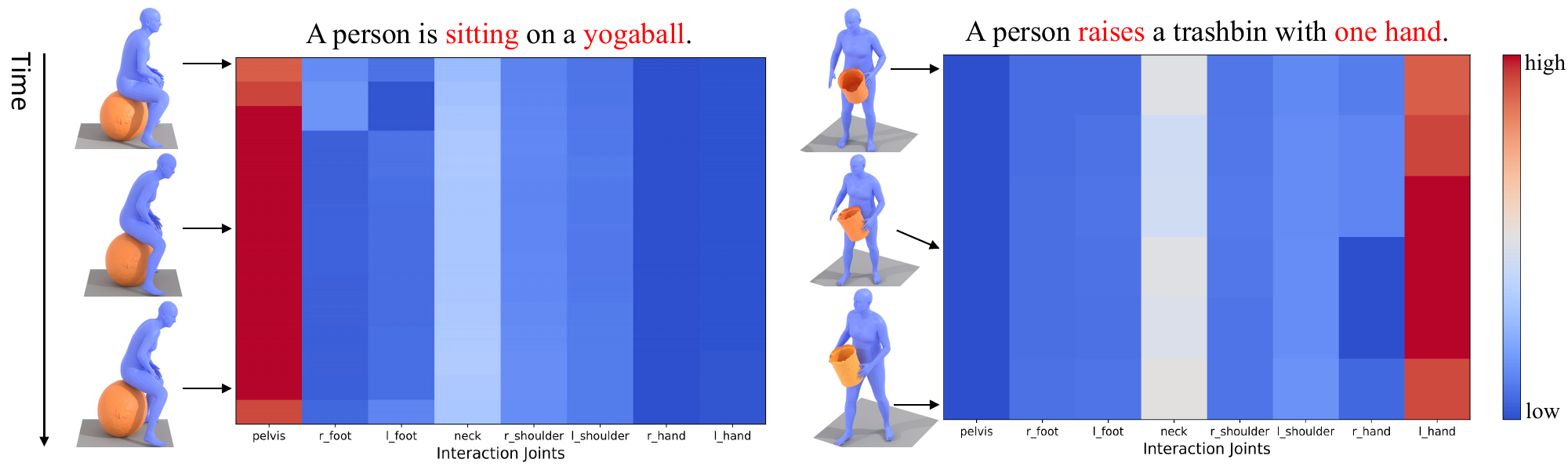}
    \caption{\textbf{Visualization of attention scores in our Kinematic-aware Decoder.} We present two examples to demonstrate that our ChainHOI can adaptively focus on the joints interacting with the target object. For other potential interaction joints that have low relevance to the target object, lower attention scores are assigned. The results show that our method effectively captures the relationship between the target object and the precise interaction joints.}
    \label{fig:weights}
\end{figure*}

\section{Failure Case Analysis} \label{sec:failure-cases}
\label{sec:failure-cases}
As shown in \cref{fig:failure-cases}, we present two typical failure cases encountered by our method:
\begin{itemize}
    \item \textit{Issues with Finger-Object Clipping}: Subfigures (a) and (b) illustrate problems related to clipping between the fingers and objects. Since the input data is represented using the SMPL model \cite{SMPL}, which does not include finger joint information, our ChainHOI is unable to accurately model the fingers, resulting in clipping between the fingers and objects.
    \item \textit{Large Contact Distances}: Subfigure (c) shows that the contact distance may be excessively large when interacting with certain complex objects. For instance, with objects such as chairs, our model struggles to learn the correct contact points and appropriate contact distances.
\end{itemize}

\section{Visualization of Attention Scores in Our Kinematic-aware Decoder. } \label{sec:visualization}

As shown in \cref{fig:weights}, we present two examples to demonstrate that our ChainHOI can adaptively focus on the joints interacting with the target object. For other potential interaction joints that have low relevance to the target object, lower attention scores are assigned. The results show that our method effectively captures the relationship between the target object and the precise interaction joints.

\section{Limitations} \label{sec: limitations}
Although our ChainHOI is capable of generating realistic and coherent human-object interactions, it still has certain limitations. First, due to the SMPL human representation used in the BEHAVE \cite{behave} and OMOMO \cite{omomo} datasets, our ChainHOI is unable to accurately model the fingers and prevent clipping between the fingers and objects. Second, as analyzed in \cref{sec:failure-cases}, our ChainHOI struggles to learn the correct contact points and appropriate contact distances for complex objects. Furthermore, due to the physical geometry information extraction method adopted in our approach, ChainHOI is unable to handle interactions between humans and non-rigid objects.

\begin{table*}[htbp] 
\centering 
\begin{tabular}{p{\textwidth}}
\toprule
    \textbf{``System Prompt:"} You are an assistant of understanding and grouping human motion instructions. Given several groups of instructions that describe interactions between humans and objects, the instructions in each group represent semantically consistent actions. Your task is to identify which groups represent semantically consistent actions. Please provide your output in the following format: [[1, 2], [3]], which represents group 1 and 2 are consistent.  \\\hline
    Here is an example: \\
    \textbf{\#\#\#Input\#\#\#:}\\
    \hspace{1em}Group 1:\\
    \hspace{2em}The person is gripping the yogamat from the front.\\
    \hspace{2em}The person has a firm grasp on the yogamat from the front.\\
    \hspace{2em}The person is clutching the yogamat from the front.\\
    \hspace{1em}Group 2: \\
    \hspace{2em}The person is clutching a yogamat against his body with his right hand.\\
    \hspace{2em}The individual is clasping a yogamat near his body with his right hand.\\
    \hspace{2em}The person is gripping a yogamat close to his body with his right hand.\\
    \hspace{1em}Group 3: \\
    \hspace{2em}A person is gripping a yogamat in front.\\
    \hspace{2em}A person is carrying a yogamat in front.\\
    \hspace{2em}A person is clutching a yogamat in front.\\
    \hspace{1em}Group 4: \\
    \hspace{2em}The individual is clutching a yogamat with his left hand, keeping it firmly against his body.\\
    \hspace{2em}A person is grasping onto a yogamat, holding it tightly against his body with his left hand.\\
    \hspace{2em}Someone holds a yogamat close to his body, with his left hand gripping onto it tightly.\\
    \hspace{1em}Group 5: \\
    \hspace{2em}The person is grasping the yogamat from the front.\\
    \hspace{2em}A person has ahold of the yogamat from the front.\\
    \hspace{2em}The person has taken possession of the yogamat from the front.\\
    \textbf{\#\#\# Output \#\#\#:} \\
    \hspace{1em} \texttt{[[1, 3, 5], [2], [4]]} \\
\hline
\end{tabular}
\caption{The prompt used to group semantically identical HOIs. } 
\label{tab:prompt_group} 
\end{table*}

\begin{table*}[htbp] 
\centering 
\begin{tabular}{p{\textwidth}}
\toprule
    \textbf{``System Prompt:"} You are an assistant responsible for evaluating and checking the consistency of human motion instructions. Given several groups of instructions that describe interactions between humans and objects, your task is to assess the semantic similarity among all groups. The instructions in one group are semantically consistent. You should output a similarity score between 0 and 1, where 1 indicates all groups are semantic similar, and 0 indicates complete inconsistency. Only when groups show completely inconsistent semantics and you are very certain, score 0. Do not output any content other than scores. \\\hline
    Here is an example: \\
    \textbf{\#\#\#\#\#\#Example 1\#\#\#\#\#\#\#\#\#}\\
    \textbf{\#\#\#Input\#\#\#}\\
    \hspace{1em} Group 1:\\
    \hspace{2em} The person is pushing the chairwood back and forth.\\
    \hspace{2em} The person is moving the chairwood back and forth. \\
    \hspace{2em} The person is exerting force on the chairwood, moving it back and forth. \\
    \hspace{1em}Group 2:\\
    \hspace{2em} A person is sitting on the chairwood. \\
    \hspace{2em} A person is occupying the chairwood. \\
    \hspace{2em} A person is positioned on the chairwood. \\
    \textbf{\#\#\#Output\#\#\#}  \\
    \hspace{1em} 0\\
    
    \textbf{\#\#\#\#\#\#Example 1\#\#\#\#\#\#\#\#\#}\\
    \textbf{\#\#\#Input\#\#\#}\\
    \hspace{1em}Group 1:\\
    \hspace{2em} The person is propelling the tablesquare with his foot. \\
    \hspace{2em} The person is nudging the tablesquare using his foot. \\
    \hspace{2em} The person is pressing the tablesquare forward by his foot. \\
    \hspace{1em}Group 2:\\
    \hspace{2em} A person is nudging the tablesquare with his foot. \\
    \hspace{2em} A person is shoving the tablesquare with his foot. \\
    \hspace{2em} A person is prodding the tablesquare with his foot. \\
    \textbf{\#\#\#Output\#\#\#}  \\
    \hspace{1em}1\\
    
    \textbf{\#\#\#\#\#\#Example 1\#\#\#\#\#\#\#\#\#}\\
    \textbf{\#\#\#Input\#\#\#}\\
    \hspace{1em}Group 1:\\
    \hspace{2em} A person clutches the boxlarge from the front. \\
    \hspace{2em} A person firmly grasps the boxlarge from the front. \\
    \hspace{2em} A person tightly holds the boxlarge from the front. \\
    \hspace{1em}Group 2:\\
    \hspace{2em} A person is grasping the boxlarge using only his left hand. \\
    \hspace{2em} A person has lifted the boxlarge using only his left hand. \\
    \hspace{2em} A person is clutching the boxlarge using only his left hand. \\
    \textbf{\#\#\#Output\#\#\#}  \\
    \hspace{1em}1\\
\hline
\end{tabular}
\caption{The prompt used to evaluate group labels to ensure consistency within each instruction group. } 
\label{tab:prompt_check} 
\end{table*}


\end{document}